\documentclass[runningheads]{llncs}
\usepackage{graphicx}
\usepackage{amsmath,amssymb} 
\usepackage{comment}
\usepackage{subcaption}
\usepackage{color}
\usepackage{longtable}
\usepackage[percent]{overpic}
\usepackage[width=122mm,left=12mm,paperwidth=146mm,height=193mm,top=12mm,paperheight=217mm]{geometry}
\usepackage{caption}
\usepackage{placeins}
\usepackage{multicol}

\newcommand{\tab}[1]{Table~\ref{tab:#1}}

\newcommand{\eq}[1]{(\ref{eq:#1})}

\begin{document}
\pagestyle{headings}
\mainmatter
\def\ECCV20SubNumber{4168}  

\title{Neural Point-Based Graphics} 

\titlerunning{Neural Point-Based Graphics}

\authorrunning{Kara-Ali Aliev, Artem Sevastopolsky, Maria Kolos et al.}

\author{Kara-Ali Aliev\textsuperscript{1}, Artem Sevastopolsky\textsuperscript{1,2}, Maria Kolos\textsuperscript{1,2}, Dmitry Ulyanov\textsuperscript{3},\\ Victor Lempitsky\textsuperscript{1,2}}
\institute{\textsuperscript{1}Samsung AI Center, \textsuperscript{2}Skolkovo Institute of Science and Technology, \textsuperscript{3}In3D.io}

\maketitle

\vspace{-0.3cm}

\begin{figure}[!h]
  \centering
  \includegraphics[trim={0 15cm 0 5cm}, clip, width=1.0\textwidth]{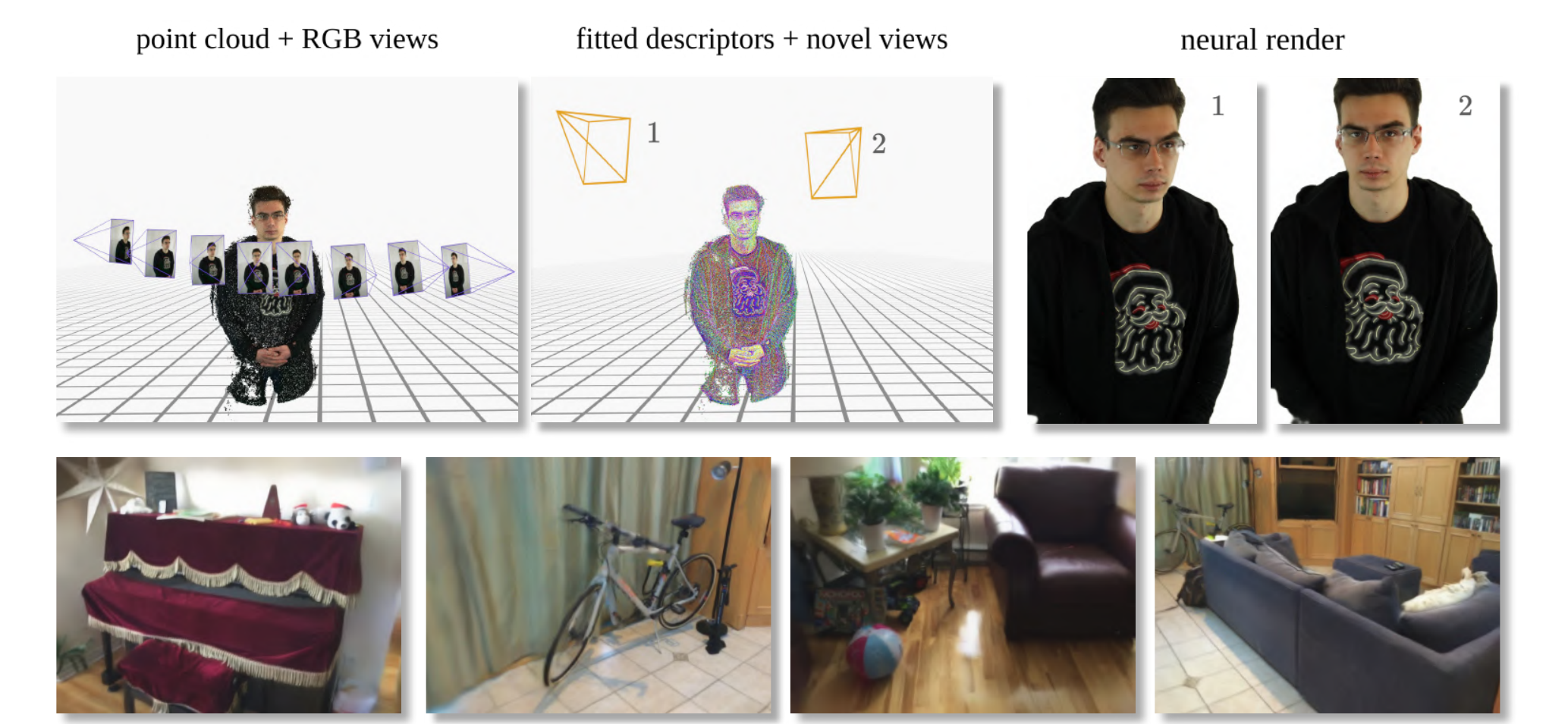}
  \captionsetup{font=scriptsize,labelfont=scriptsize}
  \caption{Given a set of RGB views and a point cloud (top-left), our approach fits a neural descriptor to each point (top-middle), after which new views of a scene can be rendered (top-right). The method works for a variety of scenes including 3D portraits (top) and interiors (bottom).}
  \label{fig:teaser}
\end{figure}

\vspace{-0.5cm}

\begin{abstract}
We present a new point-based approach for modeling the appearance of real scenes. The approach uses a raw point cloud as the geometric representation of a scene, and augments each point with a learnable neural descriptor that encodes local geometry and appearance. A deep rendering network is learned in parallel with the descriptors, so that new views of the scene can be obtained by passing the rasterizations of a point cloud from new viewpoints through this network. The input rasterizations use the learned descriptors as point pseudo-colors. We show that the proposed approach can be used for modeling complex scenes and obtaining their photorealistic views, while avoiding explicit surface estimation and meshing. In particular, compelling results are obtained for scene scanned using hand-held commodity RGB-D sensors as well as standard RGB cameras even in the presence of objects that are challenging for standard mesh-based modeling.
\end{abstract}

\section{Introduction}
\label{sect:intro}

Creating virtual models of real scenes usually involves a lengthy pipeline of operations. Such modeling usually starts with a scanning process, where the photometric properties are captured using camera images and the raw scene geometry is captured using depth scanners or dense stereo matching. The latter process usually provides noisy and incomplete point cloud that needs to be further processed by applying certain surface reconstruction and meshing approaches. Given the mesh,  the texturing and material estimation processes determine the photometric properties of surface fragments and store them in the form of 2D parameterized maps, such as texture maps~\cite{Blinn76}, bump maps~\cite{Blinn78}, view-dependent textures~\cite{Debevec98}, surface lightfields~\cite{Wood00}. Finally, generating photorealistic views of the modeled scene involves computationally-heavy rendering process such as ray tracing and/or radiance transfer estimation.

The outlined pipeline has been developed and polished by the computer graphics researchers and practitioners for decades. Under controlled settings, this pipeline yields highly realistic results. Yet several of its stages (and, consequently, the entire pipeline) remain brittle. Multiple streams of work aim to simplify the entire pipeline by eliminating some of its stages. Thus, image-based rendering techniques~\cite{McMillan95,Seitz96,Gortler96,Levoy96} aim to obtain photorealistic views by warping the original camera images using certain (oftentimes very coarse) approximations of scene geometry. Alternatively, point-based graphics~\cite{Levoy85,Grossman98,Gross02,Kobbelt04} discards the estimation of the surface mesh and use a collection of points or unconnected disks (surfels) to model the geometry. More recently, deep rendering approaches~\cite{Isola17,Nalbach17,Chen18,Bui18,Hedman18} aim to replace physics-based rendering with a generative neural network, so that some of the mistakes of the modeling pipeline can be rectified by the rendering network.

Here, we present a system that eliminates many of the steps of the classical pipeline. It combines the ideas of image-based rendering, point-based graphics, and neural rendering into a simple approach. The approach uses the raw point-cloud as a scene geometry representation, thus eliminating the need for surface estimation and meshing. Similarly to other neural rendering approaches, it also uses a deep convolutional neural network to generate photorealistic renderings from new viewpoints. The realism of the rendering is facilitated by the estimation of latent vectors (neural descriptors) that describe both the geometric and the photometric properties of the data. These descriptors are learned directly from data, and such learning happens in coordination with the learning of the rendering network (see Fig. \ref{fig:method}).

We show that our approach is capable of modeling and rendering scenes that are captured by hand-held RGBD cameras as well as simple RGB streams (from which point clouds are reconstructed via structure-from-motion or similar techniques). A number of comparisons are performed with ablations and competing approaches, demonstrating the capabilities, advantages, and limitations of the new method. In general, our results suggest that given the power of modern deep networks, the simplest 3D primitives (i.e.\ 3D points) might represent sufficient and most suitable geometric proxies for neural rendering in many cases.

 \section{Related work}
\label{sect:related}

Our approach brings together several lines of works from computer graphics, computer vision, and deep learning communities, of which only a small subset can be reviewed due to space limitations.

\paragraph{Point-based graphics.} Using points as the modeling primitives for rendering (point-based graphics) was proposed in~\cite{Levoy85,Grossman98} and have been in active development in the 2000s~\cite{Pfister00,Zwicker01,Gross02,Kobbelt04}. The best results are obtained when each point is replaced with an oriented flat circular disk (a surfel), whereas the orientations and the radii of such disks can be estimated from the point cloud data. Multiple overlapping surfels are then rasterized and linearly combined using splatting operation~\cite{Pfister00}. More recently, \cite{Bui18} has proposed to replace linear splatting with deep convolutional network. Similarly, a rendering network is used to turn point cloud rasterizations into realistic views by \cite{Meshry19}, which rasterizes each point using its color, depth, and its semantic label. Alternatively, \cite{Pittaluga19} uses a relatively sparse point cloud such as obtained by structure-and-motion reconstruction, and rasterizes the color and the high-dimensional SIFT~\cite{Lowe04} descriptor for each point.

In our work, we follow the point-based graphics paradigm as we represent the geometry of a scene using its point cloud. However, we do not use the surface orientation, or suitable disk radii, or, in fact, even color, explicitly during rasterization. Instead, we keep a 3D point as our modeling primitive and encode all local parameters of the surface (both photometric and geometric) within neural descriptors that are learned from data. We compare this strategy with the approach of \cite{Meshry19} in the experiments.

\paragraph{Deep image based rendering.} Recent years have also seen active convergence of image-based rendering and deep learning. A number of works combine warping of preexisting photographs and the use of neural networks to combine warped images and/or to post-process the warping result. The warping can be estimated by stereo matching \cite{Flynn16}. Estimating warping fields from a single input image and a low-dimensional parameter specifying a certain motion from a low-parametric family is also possible \cite{Ganin16,Zhou16}. Other works perform warping using coarse mesh geometry, which can be obtained through multi-view stereo \cite{Hedman18,Thies18} or volumetric RGBD fusion \cite{Martin18}. Alternatively, some methods avoid explicit warping and instead use some form of plenoptic function estimation and parameterization using neural networks.  Thus, \cite{Chen18} proposes network-parameterized deep version of surface lightfields. The approach~\cite{Sitzmann19} learns neural parameterization of the plenoptic function in the form of low-dimensional descriptors situated at the nodes of a regular voxel grid and a rendering function that turns the reprojection of such desriptors to the new view into an RGB image. 

Arguably most related to ours is the deferred neural rendering (DNR) system ~\cite{Thies19}. They propose to learn \textit{neural textures} encoding the point plenoptic function at different surface points alongside the neural rendering convolutional network. Our approach is similar to \cite{Thies19}, as it also learns neural descriptors of surface elements jointly with the rendering network. The difference is that our approach uses point-based geometry representation and thus avoids the need for surface estimation and meshing. We perform extensive comparison to \cite{Thies19}, and discuss relative pros and cons of the two approaches.
 \section{Methods}
\label{sect:methods}
\def\P{\mathbf{P}}
\def\D{\mathbf{D}}
\def\C{C}
\def\Loss{\mathcal{L}}
\def\s{\mathbf{s}}
\def\Rend{\mathcal{R}}
\def\I{\mathbf{I}}

Below, we explain the details of our system. First, we explain how the rendering of a new view is performed given a point cloud with learned neural descriptors and a learned rendering network. Afterwards, we discuss the process that creates a neural model of a new scene.

\begin{figure*}
  \centering
  \includegraphics[width=1\textwidth]{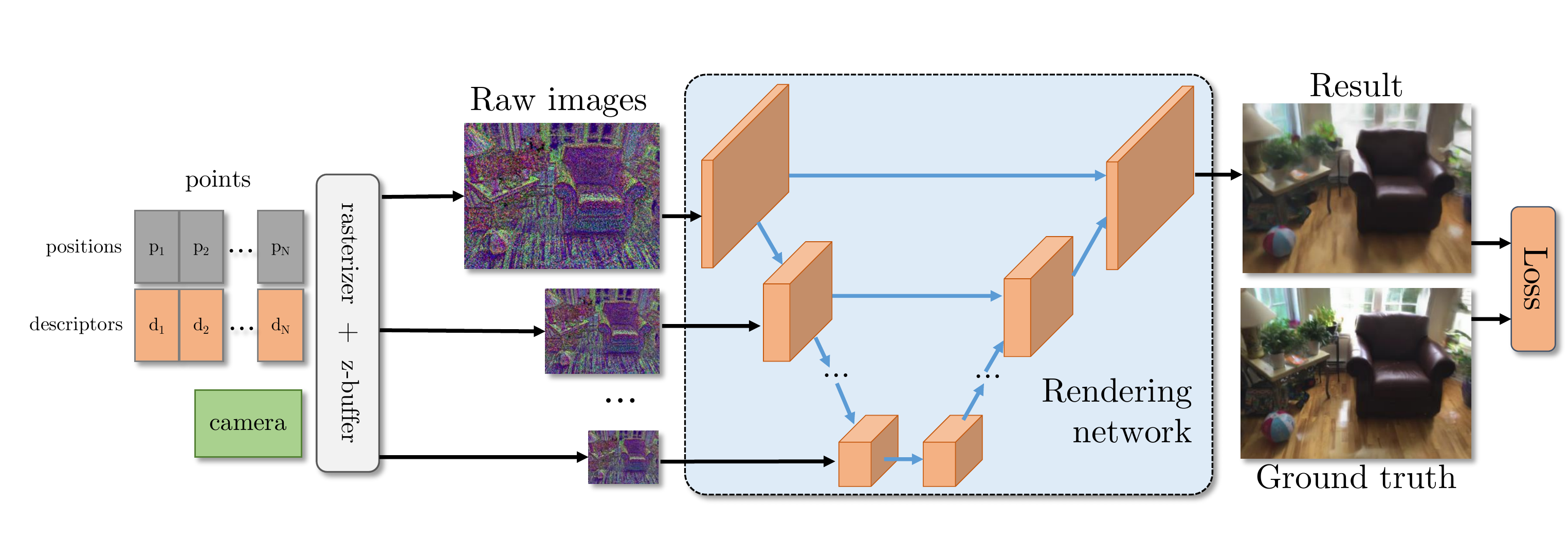}
  \caption{ An overview of our system. Given the point cloud $\P$ with neural descriptors $\D$ and camera parameters $\C$, we rasterize the points with z-buffer at several resolutions, using descriptors as pseudo-colors. We  then pass the rasterizations through the U-net-like rendering network to obtain the resulting image. Our model is fit to new scene(s) by optimizing the parameters of the rendering network and the neural descriptors by backpropagating the perceptual loss function.}
  \label{fig:method}
\end{figure*}

\subsection{Rendering}

Assume that a point cloud $\P = \{p_1,p_2, \dots,\,p_N\}$ with $M$-dimensional neural descriptors attached to each point $\D = \{d_1,d_2,\dots,d_N\}$ is given, and its rendering from a new view characterized by a camera $\C$ (including both extrinsic and intrinsic parameters) needs to be obtained. In particular, assume that the target image has $W\times{}H$-sized pixel grid, and that its viewpoint is located in point $p_0$.

The rendering process first projects the points onto the target view, using descriptors as pseudo-colors, and then uses the rendering network to transform the pseudo-color image into a photorealistic RGB image. More formally, we create an $M$-channel \textit{raw image} $S(\P,\D,\C)$ of size $W\times{}H$, and for each point $p_i$ which projects to $(x, y)$ we set $S(\P,\D,\C)[[x],[y]] = d_i$ (where $[a]$ denotes a nearest integer of $a \in \mathbb{R}$). As many points may project onto the same pixel, we use z-buffer to remove occluded points. The lack of topological information in the point cloud, however, results in hole-prone representation, such that the points from the occluded surfaces and the background can be seen through the front surface (so-called \textit{bleeding} problem). This issue is traditionally addressed through splatting, i.e.\ replacing each point with a 3D disk with a radius to be estimated from data and projecting the resulting elliptic footprint of the point onto an image. We have proposed an alternative rendering scheme that does not rely on the choice of the disk radius.

{\bf Progressive rendering.} Rather than performing splatting, we resort to multi-scale (progressive) rendering. We thus render a point cloud $T$ times onto a pyramid of canvases of different spatial resolutions. In particular, we obtain a sequence of images $S[1],S[2]\dots S[T]$, where the $i$-th image has the size of $\frac{W}{2^t}{\times}\frac{H}{2^t}$, by performing a simple point cloud projection described above. 
As a result, the highest resolution raw image $S[1]$ contains the largest amount of details, but also suffers from strong surface bleeding. The lowest resolution image $S[T]$ has coarse geometric detailization, but has the least surface bleeding, while the intermediate raw images $S[2],\dots,S[T{-}1]$ achieve different detailization-bleeding tradeoffs. 

Finally, we use a \textit{rendering network} $\Rend_\theta$ with learnable parameters $\theta$ to map all the raw images into a three-channel RGB image $I$:
\begin{equation} \label{eq:render}
    I(\P,\D,\C,\theta) = \Rend_\theta(\,S[1](\P,\D,\C),\, \dots,\, S[T](\P,\D,\C) \,)\,.
\end{equation}
The rendering network in our case is based on a popular convolutional U-Net architecture~\cite{Ronneberger15}  with gated convolutions~\cite{Yu18} for better handling of a potentially sparse input. Compared to the traditional U-Net, the rendering network architecture is augmented to integrate the information from all raw images (see Fig.~\ref{fig:method}). In particular, the encoder part of the U-Net contains several downsampling layers interleaved with convolutions and non-linearities. We then concatenate the raw image $S[i]$ to the first block of the U-Net encoder at the respective resolution. Such progressive (coarse-to-fine) mechanism is reminiscent to texture mipmapping \cite{Williams83} as well as many other coarse-to-fine/varying level of details rendering algorithms in computer graphics. In our case, the rendering network provides the mechanism for implicit level of detail selection. 

The rasterization of images $S[1],\dots,S[T]$ is implemented via OpenGL. In particular U-net network has five down- and up-sampling layers. Unless noted otherwise, we set the dimensionality of descriptors to eight ($M{=}8$).

\subsection{Model creation}
\label{subsect:learning}

We now describe the fitting process in our system. We assume that during fitting $K$ different scenes are available. For the $k$-th scene the point cloud $\P^k$ as well as the set of $L_k$ training ground truth RGB images $\I^k = \{I^{k,1},I^{k,2},\dots I^{k,L_k}\}$ with known camera parameters $\{\C^{k,1},\C^{k,2},\dots \C^{k,L_k}\}$ are given. Our fitting objective $\Loss$ then corresponds to the mismatch between the rendered and the ground truth RGB images:
\begin{align} \label{eq:learn}
    \Loss(\theta,\D^1,\D^2,\dots,\D^K) = \sum_{k=1}^K\sum_{l=1}^{L_k} \Delta\left(\Rend_\theta\left(\{ S[i](\P^k,\D^k,\C^{k,l}) \}_{i=1}^T)\,\right),\;I^{k,l}\right)\,,
\end{align}
where $\D^k$ denotes the set of neural descriptors for the point cloud of the $k$-th scene, and $\Delta$ denotes the mismatch between the two images (the ground truth and the rendered one). In our implementation, we use the perceptual loss~\cite{Dosovitskiy16,Johnson16} that computes the mismatch between the activations of a pretrained VGG network~\cite{Simonyan14}. 

The fitting is performed by optimizing the loss \eq{learn} over both the parameters $\theta$ of the rendering network \textbf{and} the neural descriptors $\{\D^1,\D^2,\dots,\D^K\}$ of points in the training set of scenes. Thus, our approach learns the neural descriptors directly from data. Optimization is performed by the ADAM algorithm~\cite{Diederik14}. Note, that the neural descriptors are updated via backpropagation through \eq{render} of the loss derivatives w.r.t.\ $S(\P,\D,\C)$ onto $d_i$.

Our system is amenable for various kinds of transfer/incremental learning. Thus, while we can perform fitting on a single scene, the results for new viewpoints tend to be better when the rendering network is fitted to multiple scenes of a similar kind. In the experimental validation, unless noted otherwise, we fit the rendering network in a two stage process. We first \textit{pretrain} the rendering network on a family of scenes of a certain kind. Secondly, we fit (\textit{fine-tune}) the rendering network to a new scene. At this stage, the learning process \eq{learn} starts with zero descriptor values for the new scene and with weights of the pretrained rendering network.

\section{Experiments}

\subsubsection{Datasets.} To demonstrate the versatility of the approach, we evaluate it on several types of real scenes. Thus, we consider three sources of data for our experiments. First, we take RGBD streams from the ScanNet dataset~\cite{Dai17b} of room-scale scenes scanned with a structured-light RGBD sensor\footnote{\texttt{https://structure.io/}}. Second, we consider RGB image datasets of still standing people captured by a mirrorless camera with high resolution (the views capture roughly 180 degrees). Finally, we consider two more scenes corresponding to two objects captured by a smartphone camera. $360^{\circ}$ camera flights of two selected objects (a potted plant and a small figurine) of a different kind captured from a circle around the object.  

For all experiments, as per the two-stage learning scheme described in Sec.~\ref{subsect:learning}, we split the dataset into three parts, unless noted otherwise: \textbf{pretraining part}, \textbf{fine-tuning part}, and \textbf{holdout part}. The pretraining part contains a set of scenes, to which we fit the rendering network (alongside point descriptors). The fine-tuning part contains a subset of frames of a new scene, which are fitted by the model creation process started with pretrained weights of the rendering network. The holdout part contains additional views of the new scene that are not shown during fitting and are used to evaluate the performance.

For the ScanNet scenes, we use the provided registration data obtained with the BundleFusion~\cite{Dai17a} method. We also use the mesh geometry computed by BundleFusion in mesh-based baselines. Given the registration data, point clouds are obtained by joining together the 3D points from all RGBD frames and using volumetric subsampling (with the grid step 1 cm) resulting in the point clouds containing few million points per scene. We pretrain rendering networks on a set of 100 ScanNet scenes. In the evaluation, we use two ScanNet scenes '\textbf{Studio}' (scene 0), which has 5578 frames, and '\textbf{LivingRoom}' (scene 24), which has 3300 frames (both scenes are not from the pretraining part). In each case, we use every $100^{\text{th}}$ frame in the trajectory for holdout and, prior to the fitting, we remove 20 preceding and 20 following frames for each of the holdout frames from the fine-tuning part to ensure that holdout views are sufficiently distinct.

\begin{figure}[!h]
    \centering
    \begin{subfigure}{.135\textwidth}
        \includegraphics[width=\textwidth]{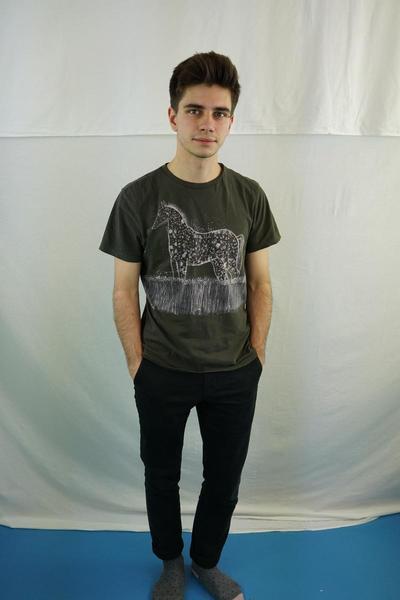}
    \end{subfigure}
    \begin{subfigure}{.135\textwidth}
        \includegraphics[width=\textwidth]{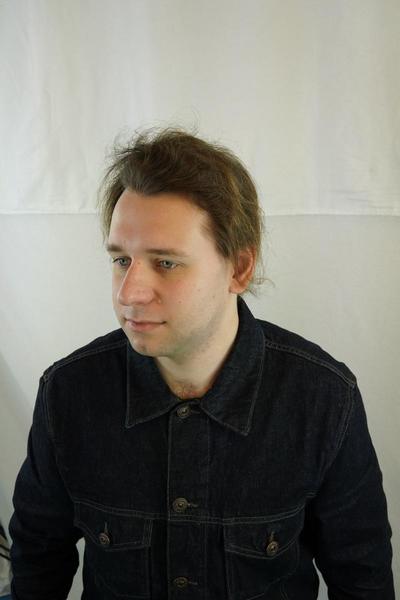}
    \end{subfigure}
    \begin{subfigure}{.135\textwidth}
        \includegraphics[width=\textwidth]{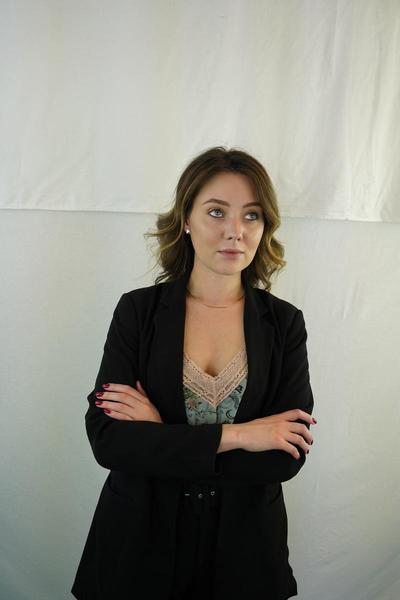}
    \end{subfigure}
    \begin{subfigure}{.135\textwidth}
        \includegraphics[width=\textwidth]{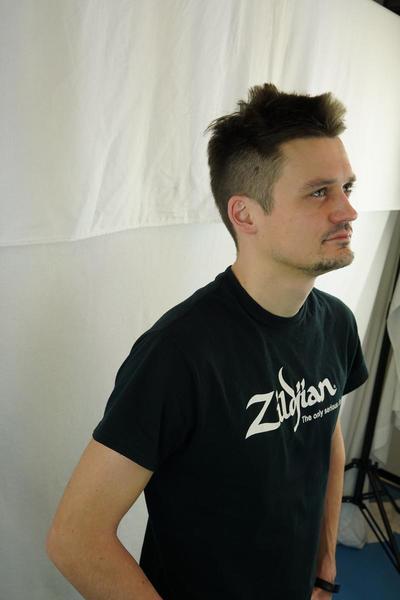}
    \end{subfigure}
    \begin{subfigure}{.135\textwidth}
        \includegraphics[width=\textwidth]{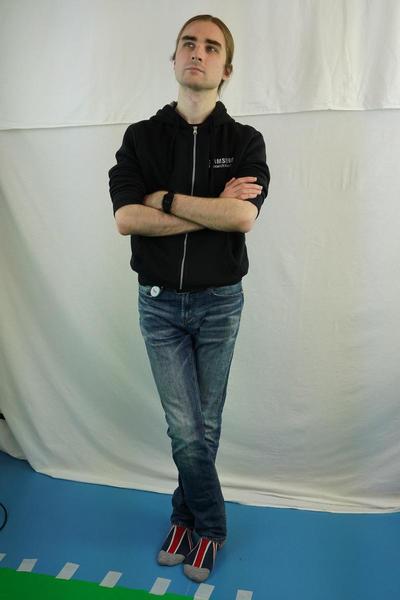}
    \end{subfigure}
    \begin{subfigure}{.135\textwidth}
        \includegraphics[width=\textwidth]{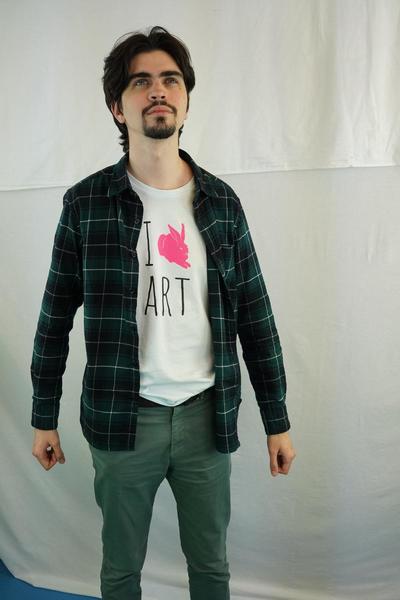}
    \end{subfigure}
    \begin{subfigure}{.135\textwidth}
        \includegraphics[width=\textwidth]{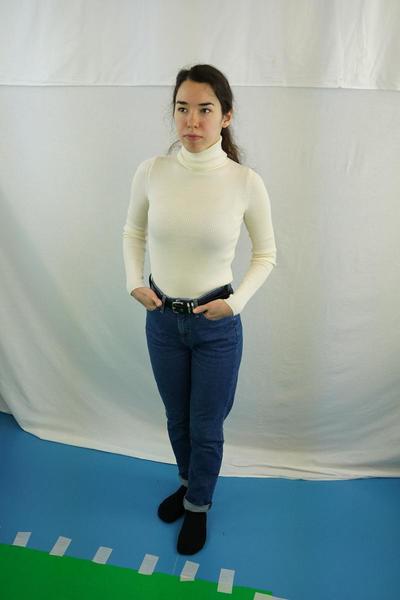}
    \end{subfigure}
    \caption{Representative samples from the People dataset used in our experiments.}
    \label{fig:people-various}
\end{figure}

For the camera-captured scenes of humans, we collected 123 sequences of 41 distinct people, each in 3 different poses, by a Sony a7-III mirrorless camera. Each person was asked to stand still (like a mannequin) against a white wall for 30--45 seconds and was photographed by a slowly moving camera along a continuous trajectory, covering the frontal half of a body, head, and hair from several angles. We then remove (whiten) the backround using the method~\cite{Gong19}, and the result is processed by the Agisoft~Metashape~\cite{agisoft} package that provides camera registration, the point cloud, and the mesh by running proprietary structure-from-motion and dense multiview stereo methods. Each sequence contains 150-200 ten megapixel frames with high amount of fine details and varying clothing and hair style (Fig.~\ref{fig:people-various}). The pretraining set has 102 sequences of 38 individuals, and three scenes of three different individuals were left for validation. Each of the validation scenes is split into fine-tuning (90\% of frames) and holdout (10\% of frames) sets randomly.

In addition, we used a smartphone (Galaxy S10e) to capture $360^{\circ}$ sequences of two scenes containing an \textbf{Owl} figurine (61 images) and a potted \textbf{Plant} (92 images). All frames of both scenes were segmented manually via tools from Adobe Photoshop CC software package. We split fine-tuning and holdout parts in the same manner as in People dataset.

\begin{figure*}[!h]
    \centering
    \captionsetup[subfigure]{labelformat=empty}
      \begin{subfigure}{0.18\linewidth}
      \includegraphics[trim={0 0 12.5cm 0}, clip,width=\linewidth]{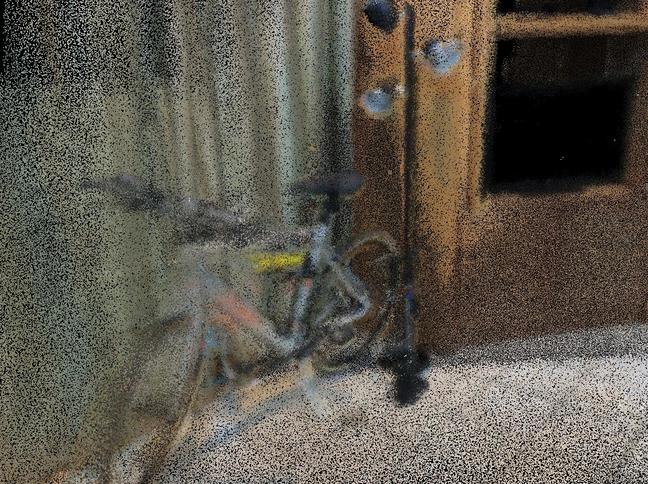}
      \end{subfigure}
      \begin{subfigure}{0.18\linewidth}
      \includegraphics[trim={0 0 12.5cm 0}, clip,width=\linewidth]{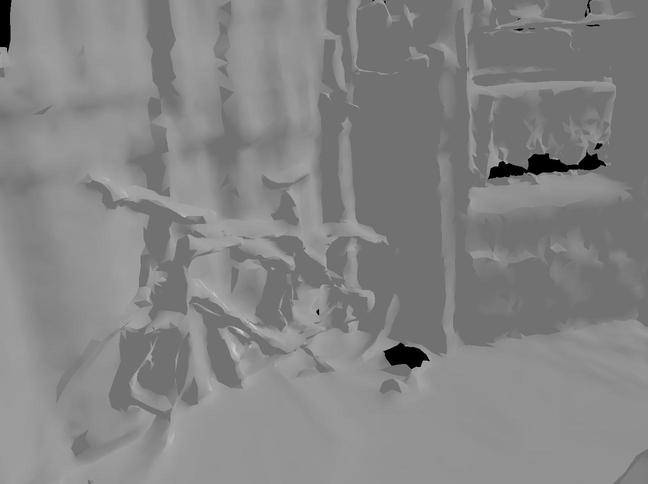}
      \end{subfigure}
      \begin{subfigure}{0.18\linewidth}
      \includegraphics[trim={0 0 12.5cm 0}, clip,width=\linewidth]{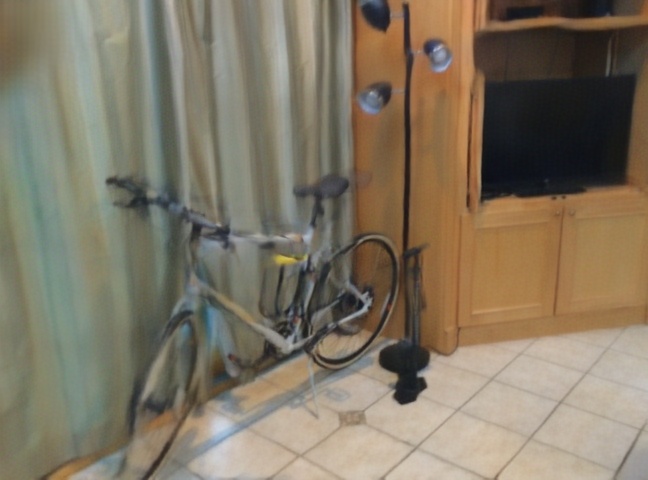}
      \end{subfigure}
      \begin{subfigure}{0.18\linewidth}
      \includegraphics[trim={0 0 12.5cm 0}, clip,width=\linewidth]{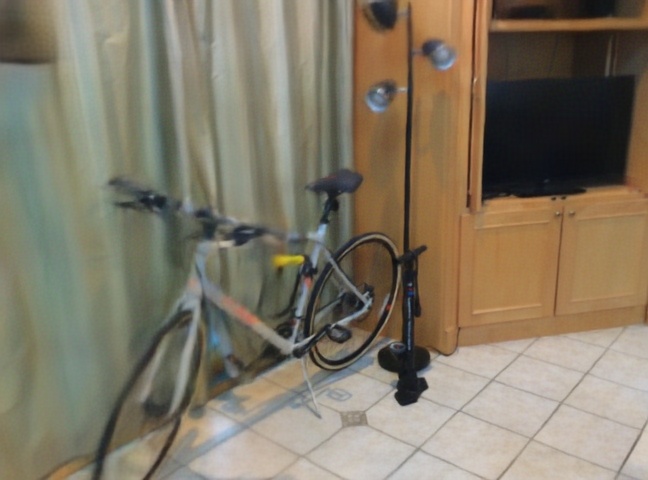}
      \end{subfigure}
      \begin{subfigure}{0.18\linewidth}
      \includegraphics[trim={0 0.25cm 9.5cm 0}, clip,width=\linewidth]{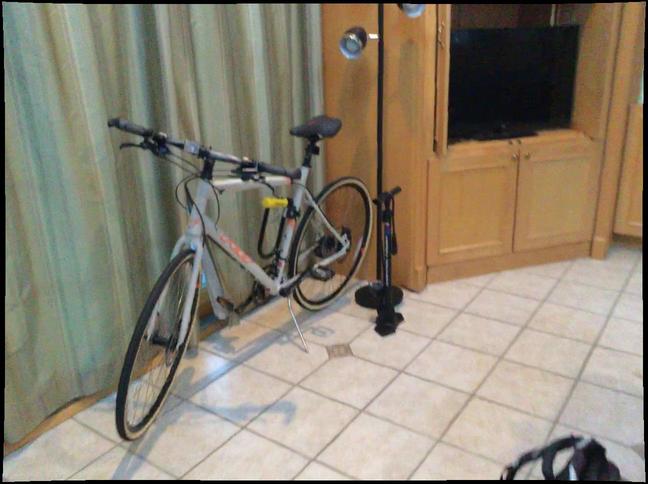}
      \end{subfigure}
      \hfill
      \begin{subfigure}{0.18\linewidth}
      \includegraphics[trim={5cm 0 7.5cm 0}, clip,width=\linewidth]{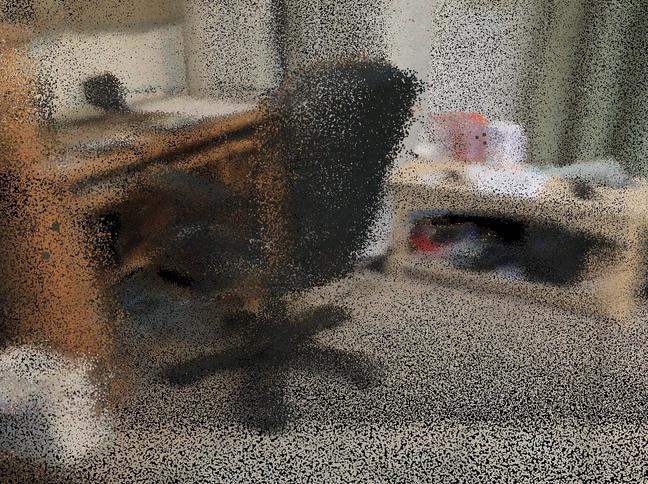}
      \caption{Point Cloud}
      \end{subfigure}
      \begin{subfigure}{0.18\linewidth}
      \includegraphics[trim={5cm 0 7.5cm 0}, clip,width=\linewidth]{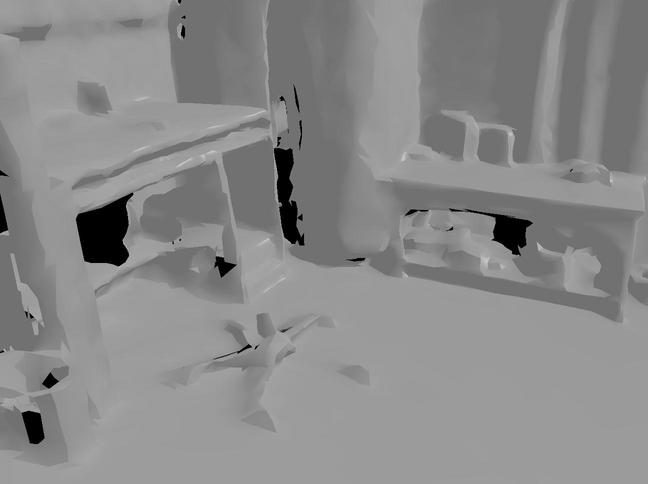}
      \caption{Mesh}
      \end{subfigure}
      \begin{subfigure}{0.18\linewidth}
      \includegraphics[trim={5cm 0 7.5cm 0}, clip,width=\linewidth]{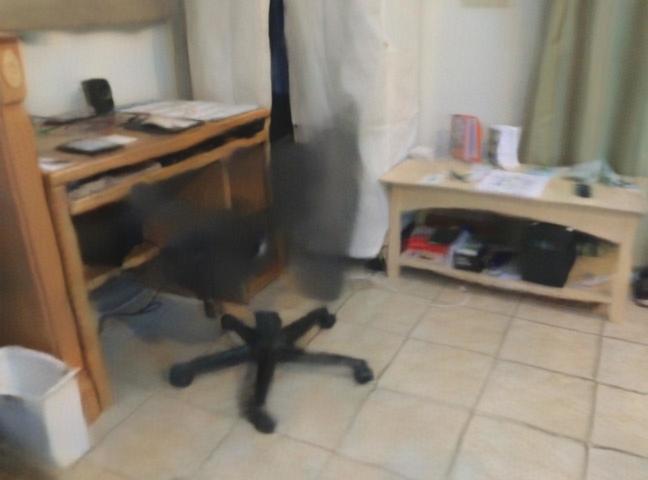}
      \caption{DNR}
      \end{subfigure}
      \begin{subfigure}{0.18\linewidth}
      \includegraphics[trim={5cm 0 7.5cm 0}, clip,width=\linewidth]{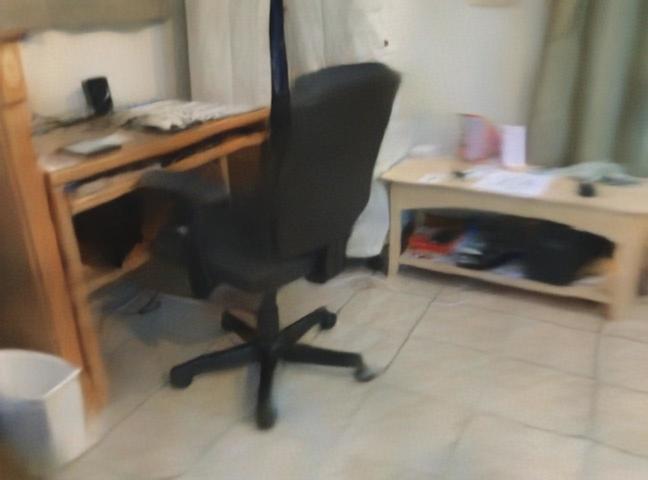}
      \caption{Ours}
      \end{subfigure}
      \begin{subfigure}{0.18\linewidth}
      \includegraphics[trim={8.5cm 0.25cm 1cm 0}, clip,width=\linewidth]{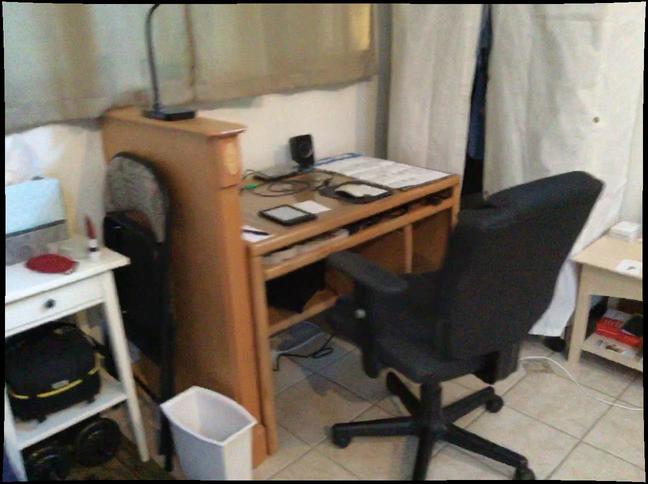}
      \caption{Nearest train}
      \end{subfigure}
    \caption{Comparative results on the 'Studio' dataset (from \cite{Dai17b}). We show the textured mesh, the colored point cloud, the results of three neural rendering systems (including ours), and the ground truth. Our system can successfully reproduce details that pose challenge for meshing, such as the wheel of the bicycle.} 
    \label{fig:studio-holdout}
\end{figure*}

\begin{figure*}[!h]
      \centering
      \begin{subfigure}[c]{0.23\textwidth}
      \centering
      \includegraphics[width=\linewidth]{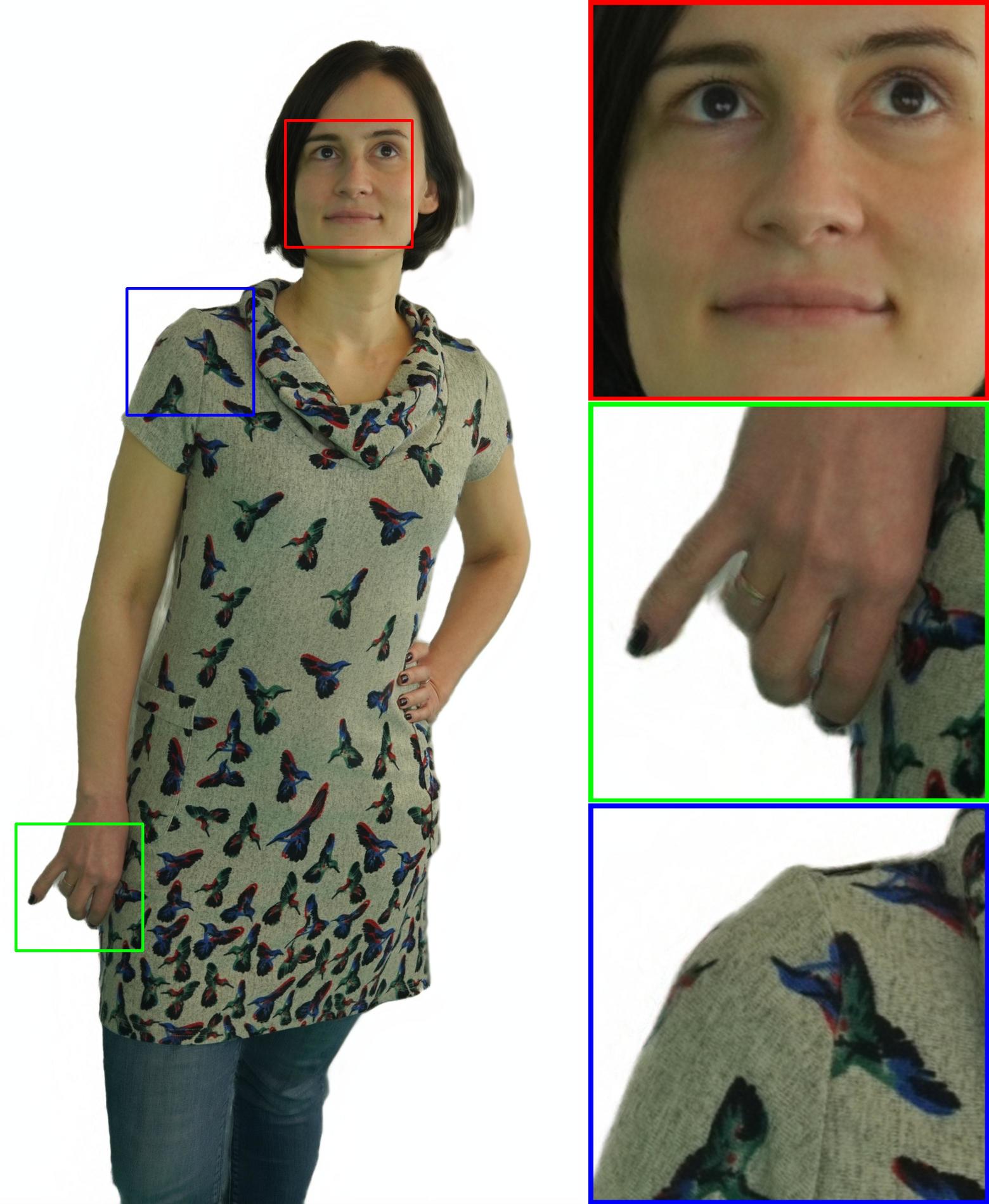}
      \end{subfigure}
       \hfill
      \begin{subfigure}[c]{0.23\textwidth}
      \centering
      \includegraphics[width=\linewidth]{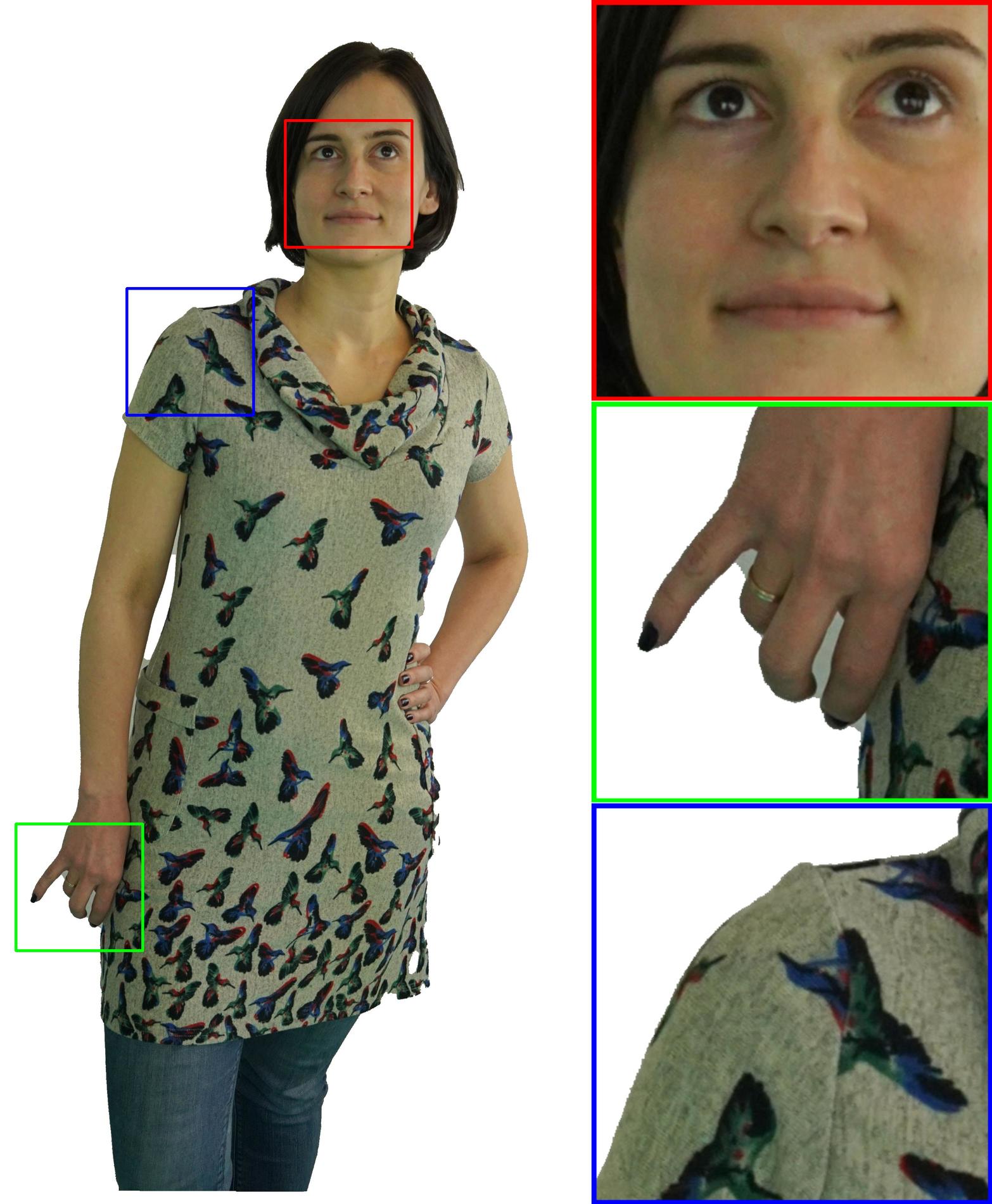}
      \end{subfigure}
       \hfill
      \begin{subfigure}[c]{0.23\textwidth}
      \centering
      \includegraphics[width=\linewidth]{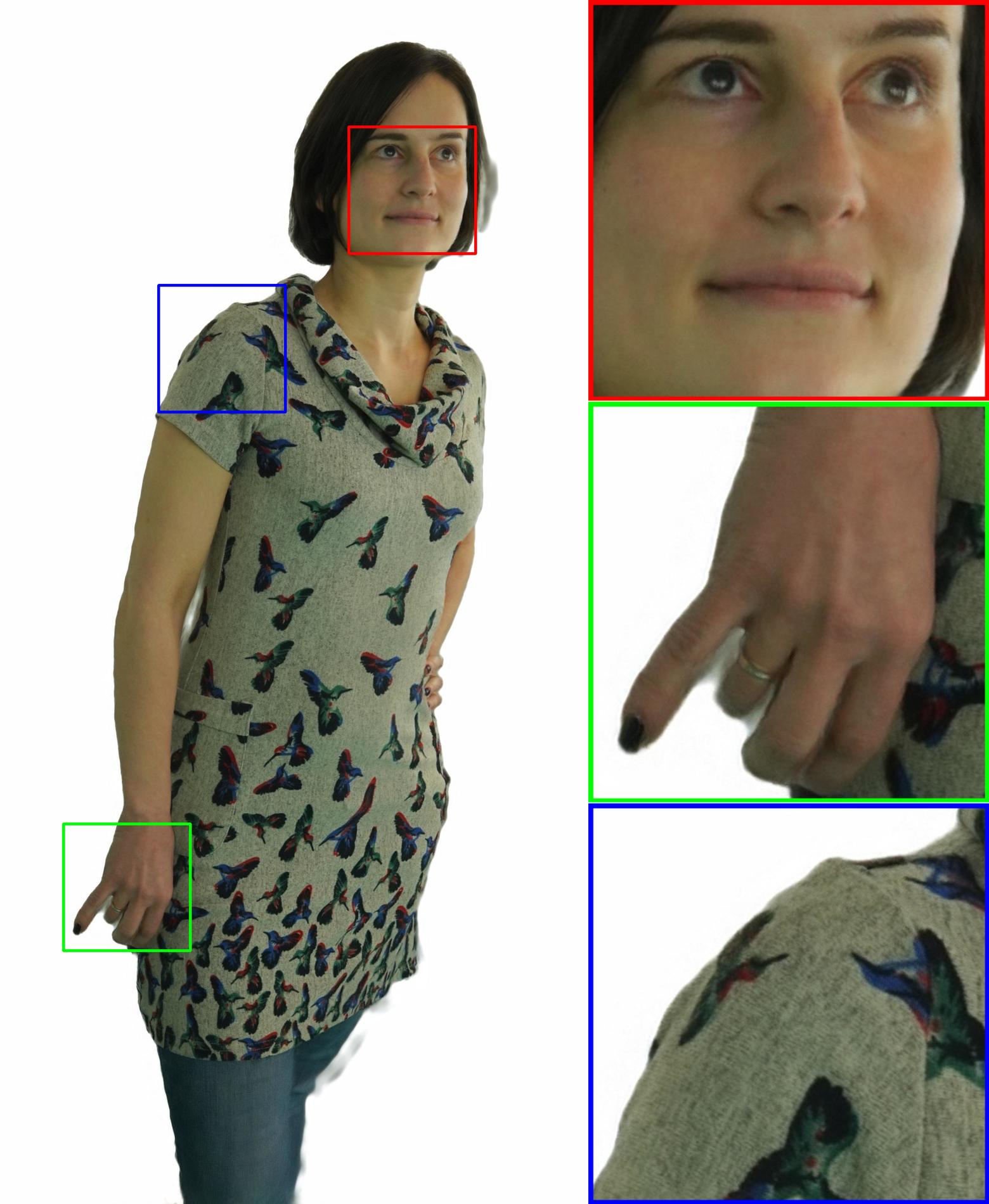}
      \end{subfigure}
      \hfill
      \begin{subfigure}[c]{0.23\textwidth}
      \centering
      \includegraphics[width=\linewidth]{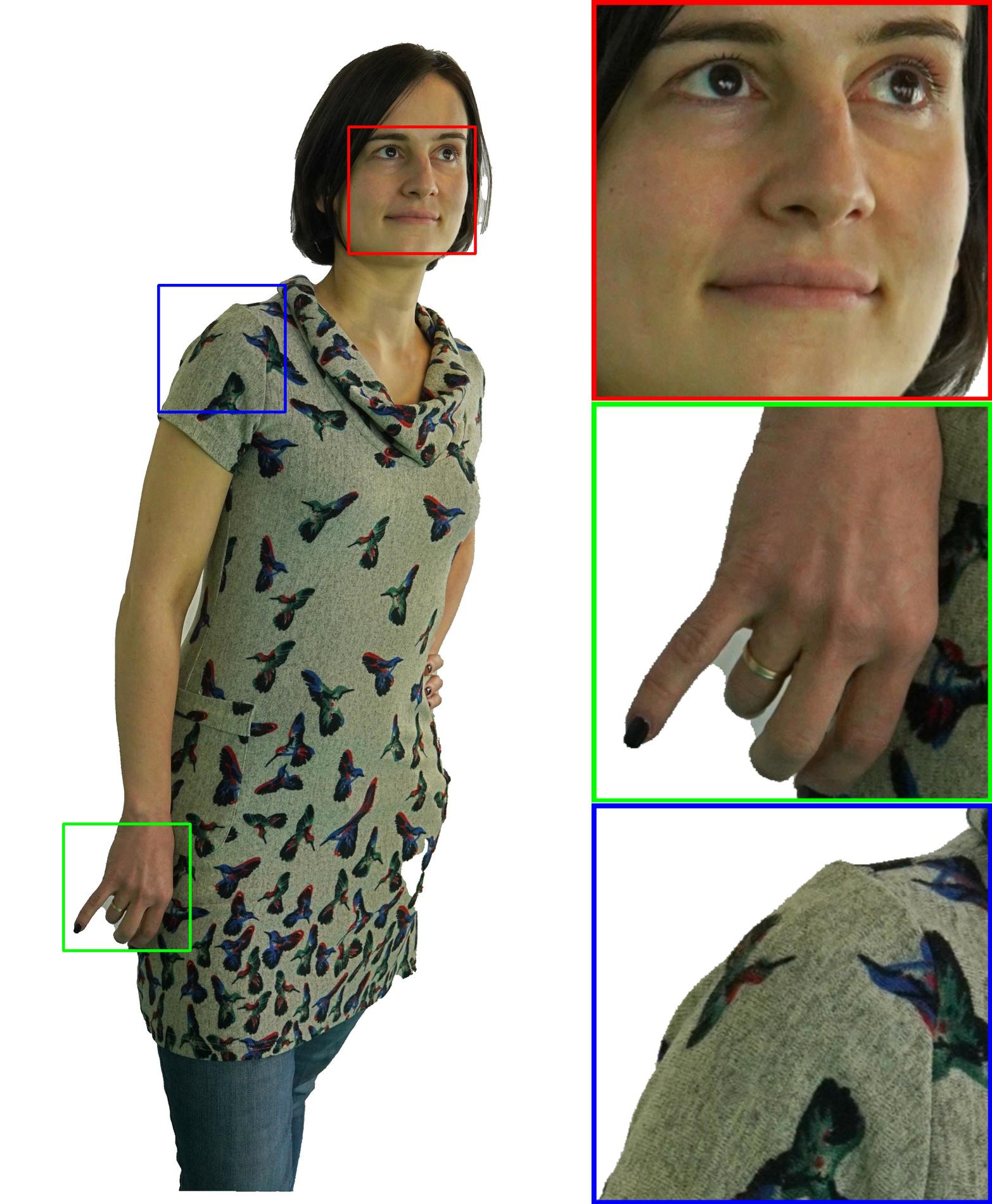}
      \end{subfigure}
      
      \medskip
      \begin{subfigure}[c]{0.23\textwidth}
      \centering
      \includegraphics[width=\linewidth]{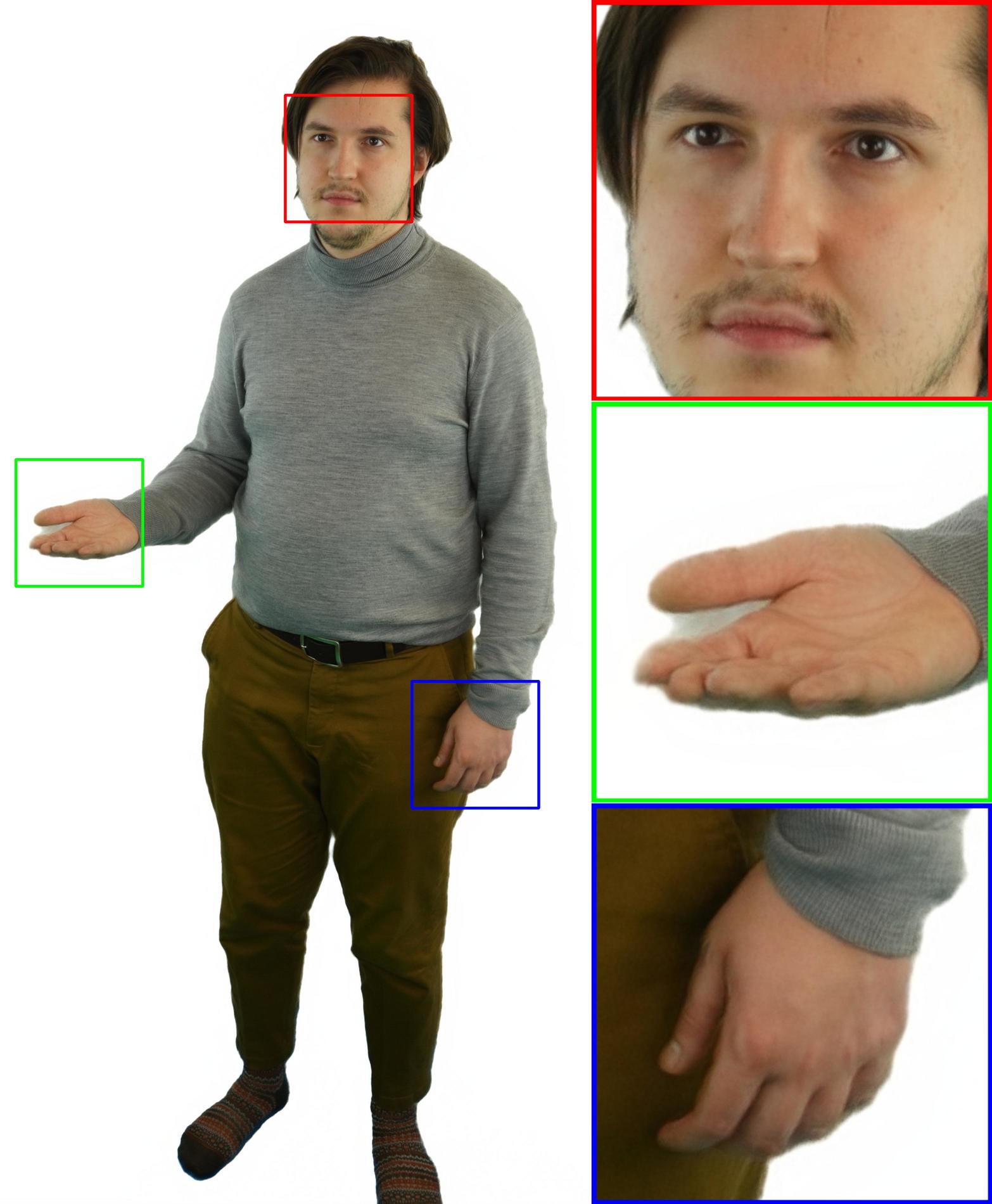}
      \caption*{Ours}
      \end{subfigure}
       \hfill
      \begin{subfigure}[c]{0.23\textwidth}
      \centering
      \includegraphics[width=\linewidth]{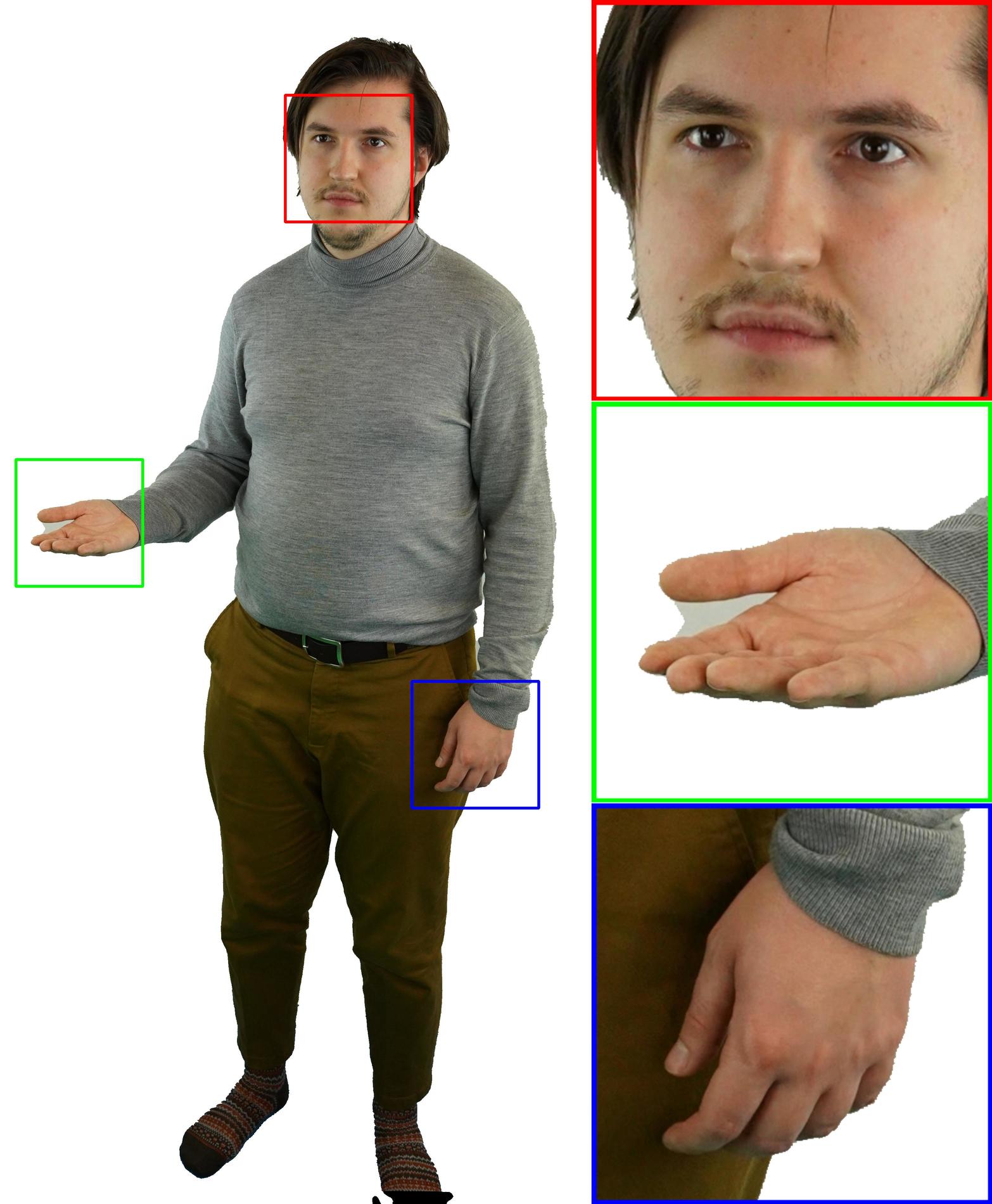}
      \caption*{Ground truth}
      \end{subfigure}
       \hfill
      \begin{subfigure}[c]{0.23\textwidth}
      \centering
      \includegraphics[width=\linewidth]{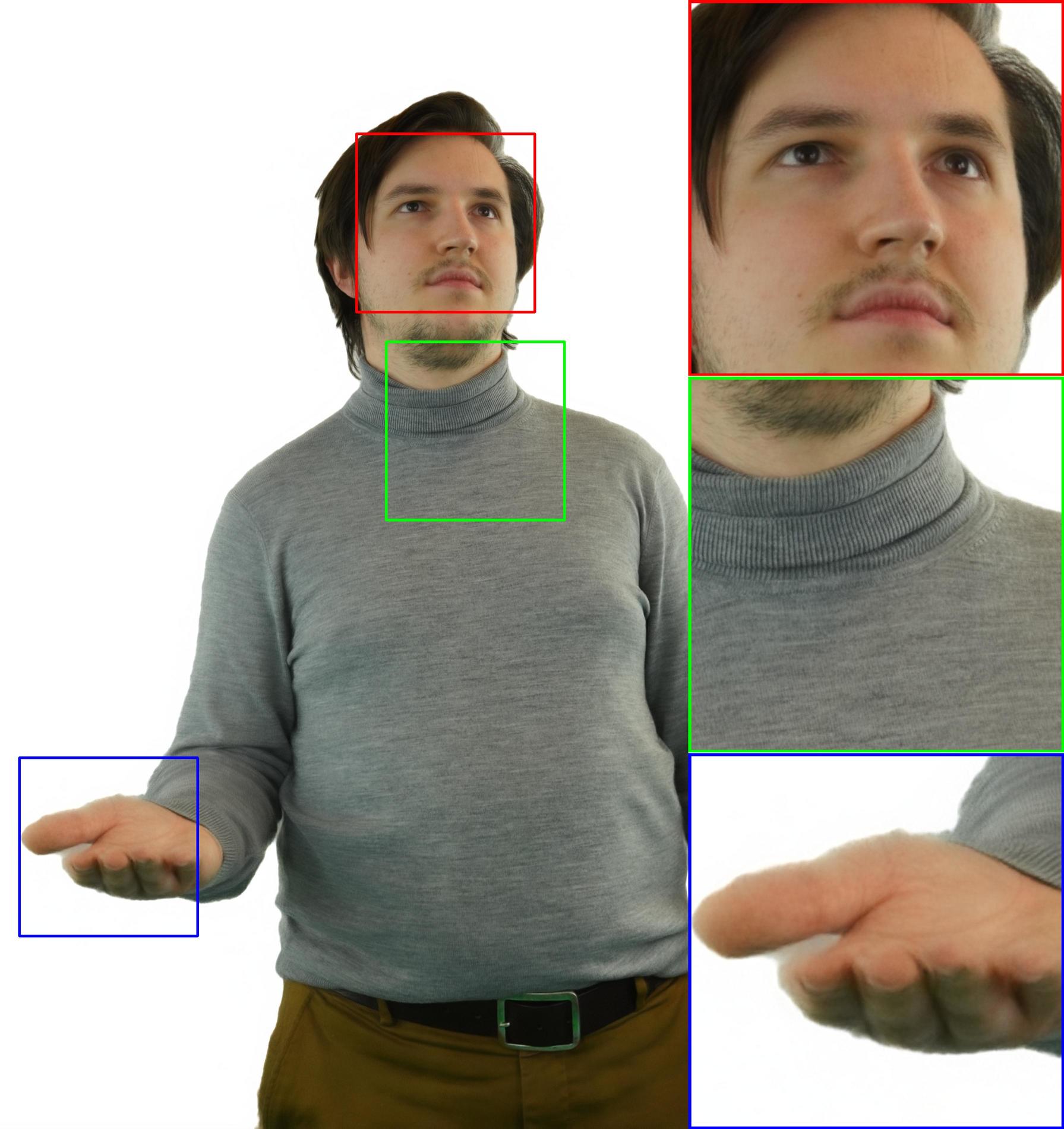}
      \caption*{Ours}
      \end{subfigure}
       \hfill
      \begin{subfigure}[c]{0.23\textwidth}
      \centering
      \includegraphics[width=\linewidth]{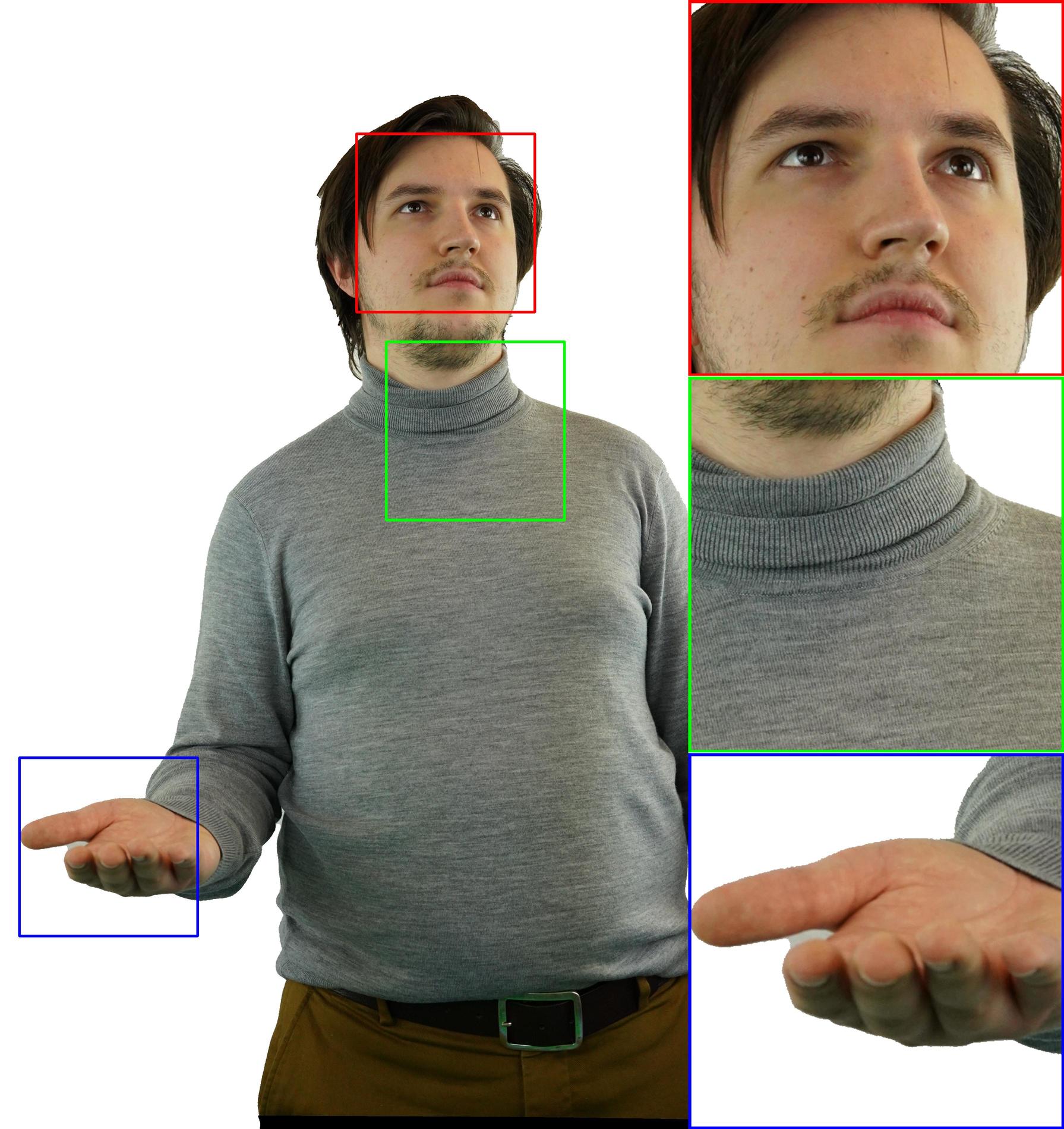}
      \caption*{Ground truth}
      \end{subfigure}
    \caption{Results on the holdout frames from the 'Person 1' and 'Person 2' scenes. Our approach successfully transfers fine details to new views.}
    \label{fig:people-holdout}
\end{figure*}

\begin{figure*}[!h]
      \rotatebox{90}{Plant}
      \begin{subfigure}{0.32\linewidth}
      \includegraphics[width=\textwidth]{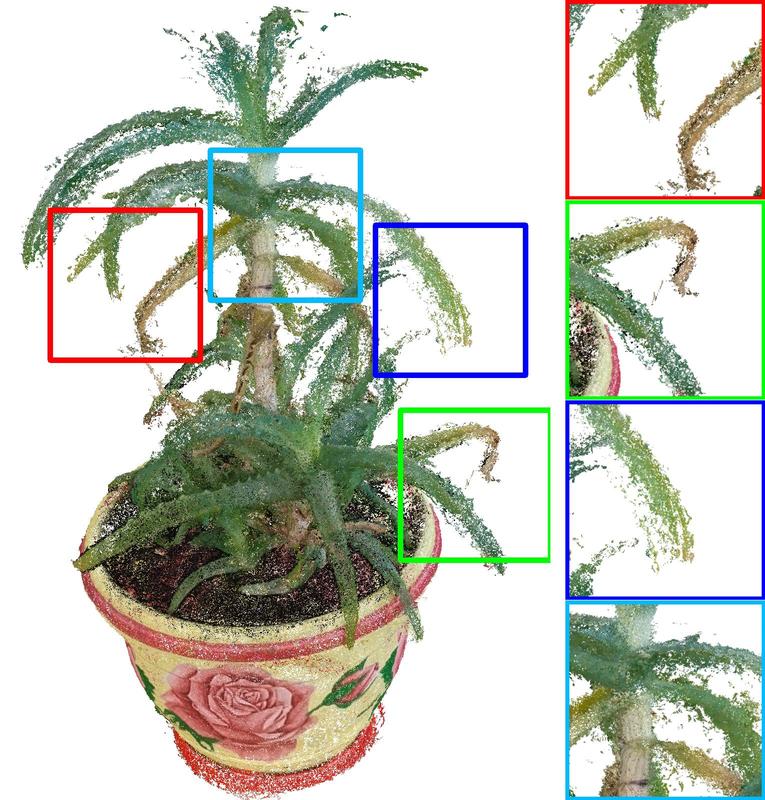}
      \caption{Point cloud}
      \end{subfigure}
      \begin{subfigure}{0.32\linewidth}
      \includegraphics[width=\textwidth]{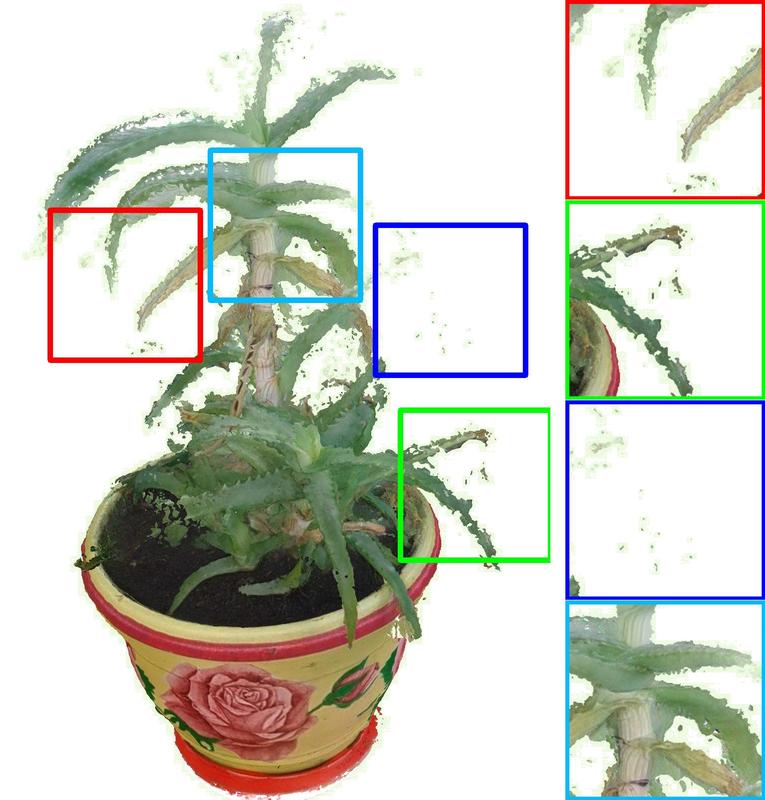}
      \caption{Mesh+Texture}
      \end{subfigure}
      \begin{subfigure}{0.32\linewidth}
      \includegraphics[width=\textwidth]{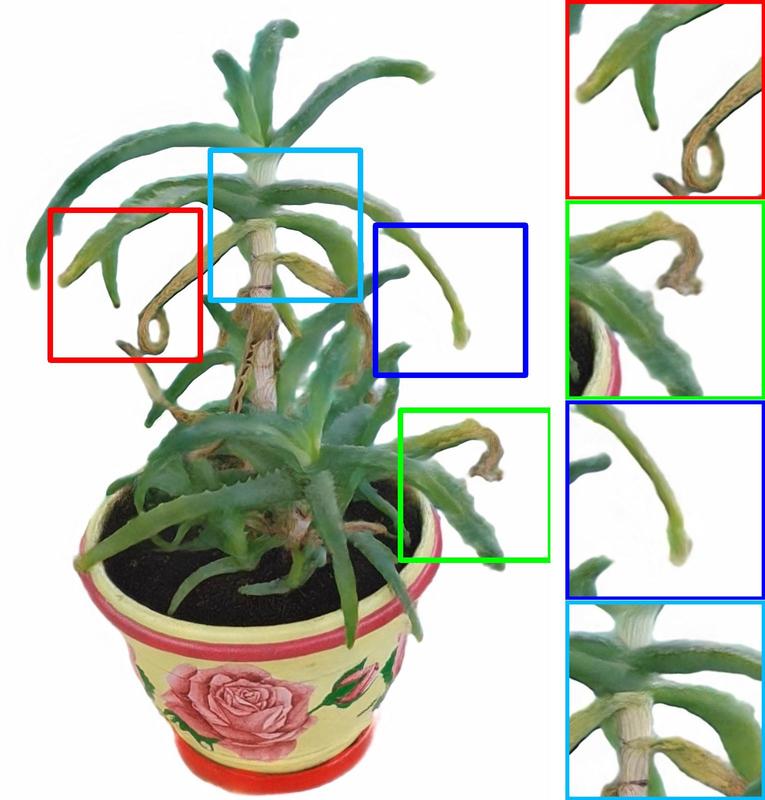}
      \caption{Pix2Pix}
      \end{subfigure}
      
      \rotatebox{90}{Plant}
      \begin{subfigure}{0.32\linewidth}
      \includegraphics[width=\textwidth]{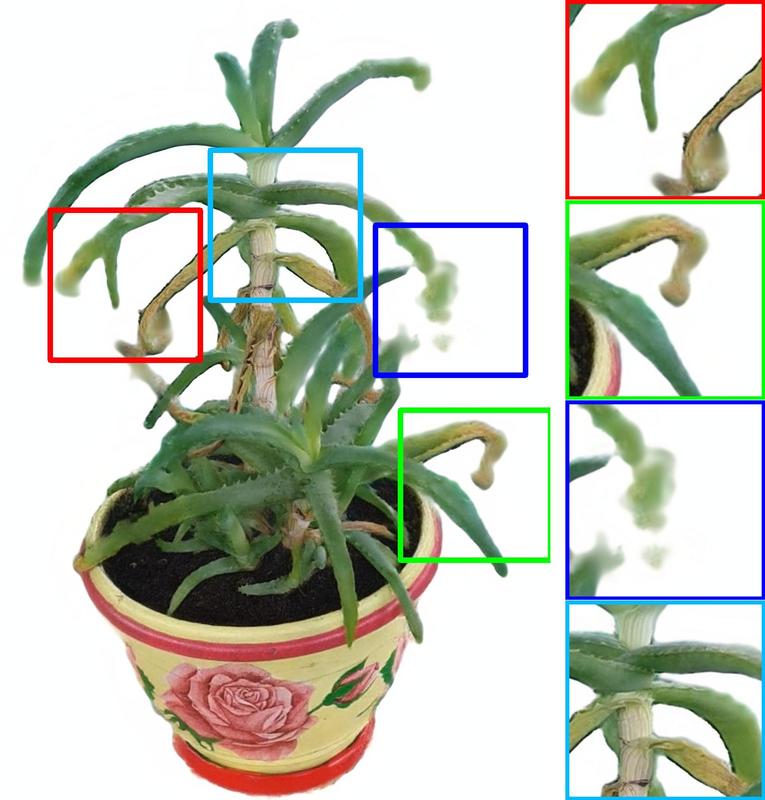}
      \caption{Deferred Neural Rendering}
      \end{subfigure}
      \begin{subfigure}{0.32\linewidth}
      \includegraphics[width=\textwidth]{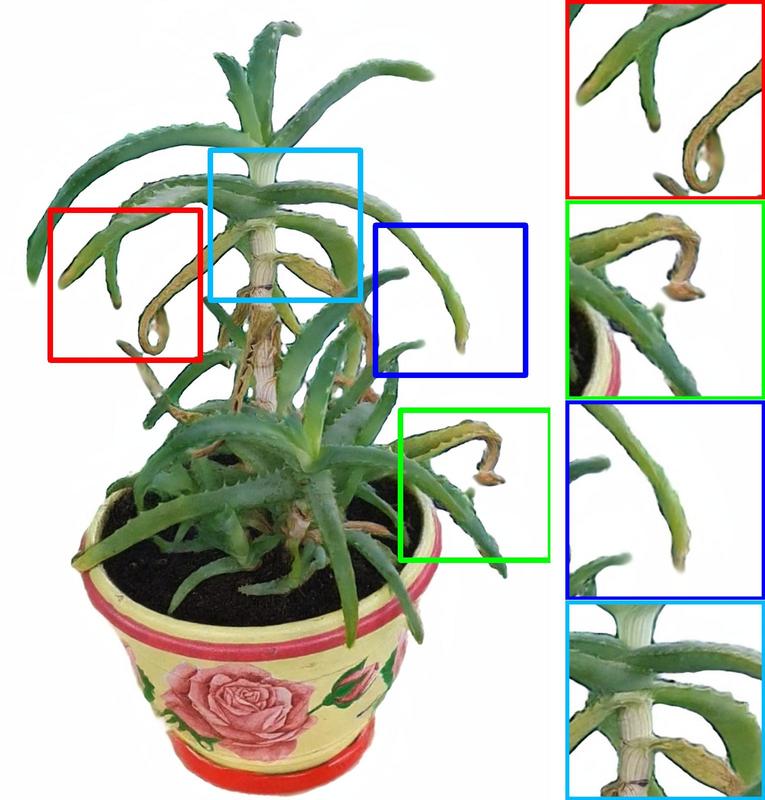}
      \caption{Ours}
      \end{subfigure}
      \begin{subfigure}{0.32\linewidth}
      \includegraphics[width=\textwidth]{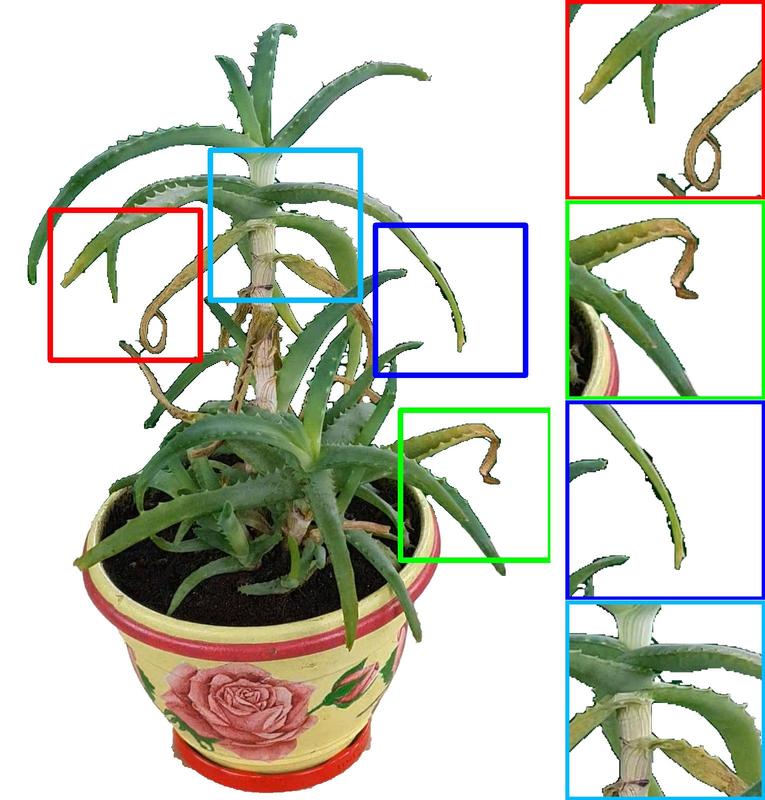}
      \caption{Ground truth}
      \end{subfigure}

    \caption{Comparative results on the holdout frame from the 'Plant' scene. Our method better preserves thin parts of the scene.}
    \label{fig:plant-holdout}
\end{figure*}

\begin{table}[!h]
\caption{Comparison with the state-of-the-art for all considered hold-out scenes from various sources: two scenes from ScanNet, two people captured by a professional camera, and two objects photographed by a smartphone. We assess all methods with respect to widely used perceptual metrics correlated with visual similarity of predicted and ground truth images (LPIPS, FID) and to VGG loss used in our experiments.}
\vspace{0.2cm}
\begin{subtable}{1\textwidth}
\begin{tabular}{r|c|c|c|c|c|c}
                            & \multicolumn{3}{c|}{ScanNet - LivingRoom}                                                       & \multicolumn{3}{c}{ScanNet - Studio}                                                           \\
Method                      & VGG $\downarrow$                 & LPIPS $\downarrow$              & FID $\downarrow$ & VGG  $\downarrow$                & LPIPS  $\downarrow$        & FID $\downarrow$ \\ \hline
Pix2Pix            & 751.04                           & 0.564                           & 192.82           & 633.30                           & 0.535                       & 127.49           \\
Pix2Pix (slow)     & 741.09                           & 0.547                           & 187.92           & 619.04                           & 0.509                      & 109.46           \\
Neural Rerendering          & 751.52                           & 0.580                           & 206.90           & 634.93                           & 0.529                      & 119.16           \\
Neural Rerendering (slow)   & 739.82                           & 0.542                           & 186.27           & 620.98                           & 0.507                      &  108.12       \\
Textured mesh               & 791.26                           & 0.535                           & 152.02            & 690.67                           & 0.540                     & 97.95           \\
Deferred Neural Rendering & \textbf{725.23}                     & 0.492                         & \textbf{129.33}   & 603.63                           & 0.484                      & 84.92  \\ \hline
Ours                        & 727.38                           & \textbf{0.488}                          & 138.87            & \textbf{595.24}                   & \textbf{0.472}    & \textbf{76.73}          
\end{tabular}
\label{tab:comparison_scannet}
\end{subtable}

\vspace{0.2cm}
\begin{subtable}{1\textwidth}
\begin{tabular}{r|c|c|c|c|c|c}
                          & \multicolumn{3}{c|}{People - Person 1}                            & \multicolumn{3}{c}{People - Person 2}                              \\
Method                    & VGG $\downarrow$ & LPIPS $\downarrow$ & FID $\downarrow$ & VGG  $\downarrow$ & LPIPS  $\downarrow$ & FID $\downarrow$ \\ \hline
Pix2Pix                   & 209.16           & 0.1016             & 51.38            & 186.89            & 0.1642              & 114.93           \\
Pix2Pix (slow)            & 204.45           & 0.0975             & 47.14            & 179.99            & 0.1566              & 102.62           \\
Textured mesh             & \textbf{155.37}  & 0.0698             & 60.62            & 163.73            & 0.1404              & 96.20            \\
Deferred Neural Rendering & 184.86           & 0.0659             & 34.41            & 163.13            & 0.1298              & 78.70            \\ \hline
Ours                      & 181.11           & \textbf{0.0602}    & \textbf{32.63}   & \textbf{161.18}   & \textbf{0.1131}     & \textbf{77.92}  
\end{tabular}
\label{tab:comparison_people}
\end{subtable}

\vspace{0.2cm}
\begin{subtable}{1\textwidth}
\begin{tabular}{r|c|c|c|c|c|c}
                          & \multicolumn{3}{c|}{Plant}                               & \multicolumn{3}{c}{Owl}                                   \\
Method                    & VGG $\downarrow$ & LPIPS $\downarrow$ & FID $\downarrow$ & VGG  $\downarrow$ & LPIPS  $\downarrow$ & FID $\downarrow$ \\ \hline
Pix2Pix          & 85.47            & 0.0443             & 52.95            & 34.30             & 0.0158              & 124.63           \\
Pix2Pix (slow)   & 82.81            & 0.0422             & 48.89            & 32.93             & 0.0141              & 101.65           \\
Textured mesh             & 101.56           & 0.0484             & 95.60            & 36.58             & 0.0145              & 141.66           \\
Deferred Neural Rendering & 77.55            & 0.0377             & 49.61            & \textbf{28.12}             & \textbf{0.0096}              & \textbf{54.14}            \\ \hline
Ours                      & \textbf{75.08}            & \textbf{0.0373}             & \textbf{41.67}            & 29.69             & 0.0103              & 78.55           
\end{tabular}
\label{tab:comparison_things}
\end{subtable}
\label{tab:comparison_with_others}
\end{table}

\subsubsection{Comparison with state-of-the-art.} We compare our method to several neural rendering approaches on the evaluation scenes. Most of these approaches have a rendering network similar to our method, which takes an intermediate representation and then is trained to output the final RGB image. Unless stated otherwise, we use the network described in Section~\ref{sect:methods}. It is lightweight with 1.96M parameters and allows us to render real-time, taking 62ms on GeForce RTX 2080 Ti to render a FullHD image. For all the approaches we use the same train time augmentations, particularly random 512x512 crops and 2x zoom-in and zoom-out.

The following methods were compared:
\begin{itemize}
    \item \textbf{Ours.} During learning, we both optimize the neural descriptors and fine-tune the rendering network on the fine-tuning part. 
    
    \item \textbf{Pix2Pix.} In this variant, we evaluate an ablation of our point-based system without neural descriptors. Here, we learn the rendering network that maps the point cloud rasterized in the same way as in our method. However, instead of neural descriptors, we use the color of the point (taken from the original RGBD scan/RGB image). The rendering network is then trained with the same loss as ours.
    
    \item \textbf{Pix2Pix (slow).} We observed that our method features neural descriptors which increases the number of parameters to be learned. For the sake of fair comparison, we therefore evaluated the variant of Pix2Pix with the rendering network with doubled number of channels in all intermediate layers (resulting in $\sim$4x parameters and FLOPs).
    
    \item \textbf{Neural Rerendering in the Wild.} Following~\cite{Meshry19}, we have augmented the input of the Pix2Pix method with the segmentation labels (one-hot format) and depth values. We have not used the appearance modeling from \cite{Meshry19}, since lightning was consistent within each dataset.
    
    \item \textbf{Neural Rerendering in the Wild (slow).} Same as previous, but twice larger number of channels in the rendering network. Due to the need to have meaningful segmentation labels, we have considered this and the previous methods only for ScanNet comparisons, where such labels are provided with the dataset.
    
    \item \textbf{Mesh+Texture.} In this baseline, given the mesh of the scene obtained with either BundleFusion or Metashape (depending on the dataset used), we learn the texture via backpropagation of the same loss as used in our method through the texture mapping process onto the texture map. This results in a ``classical'' scene representation of the textured mesh.
    
    \item \textbf{Deferred Neural Rendering (DNR).} We implemented the mesh-based approach described in ~\cite{Thies19}. As suggested, we use hierarchical neural textures with five scales (maximum $2048{\times}2048$) each having eight channels (same as the descriptor size $M$ in our method). The rendering network is then trained with the same loss as ours. Generally, this method can be seen as the analog of our method with point-based geometric proxy replaced with mesh-based proxy. 
\end{itemize}

We compare the methods on ScanNet (two scenes following pretraining on 100 other scenes), on People (two people following pretraining on 102 scenes of 38 other people), as well as on 'Owl' and 'Plant' scenes (following the pretraining on People). The quantitative results of the comparison are shown in Table~\ref{tab:comparison_with_others}. All comparisons are measured on the holdout parts, for which we compare the obtained and the ground truth RGB images. We stress that we keep the holdout viewpoints sufficiently dissimilar from the viewpoints of images used for fine-tuning. For all experiments \textit{nearest train view} is defined as follows. Given a novel view, we sort train views by angle deviation from the novel view, then leave top 5\% closest by angle and pick the view closest by distance. Angle proximity is more critical since we use zoom augmentation in training which compensate distance dissimilarity.

We report the value of the loss on the holdout part (\textit{VGG}) as well as two common metrics (Learned Perceptial Similarity -- \textit{LPIPS}~\cite{Zhang18} and Frechet Inception Distance -- \textit{FID}~\cite{Heusel17}). We also show qualitative comparisons on the holdout set frames in Figures~\ref{fig:studio-holdout}--\ref{fig:plant-holdout}, where we also show the point cloud, and renderings from completely different viewpoints in Figures~\ref{fig:far_views_people}--\ref{fig:far_views_things_scannet}. Further comparisons can be found in \textbf{Supplementary video}. 

Generally, both the quantitative and the qualitative comparison reveals the advantage of using \textit{learnable neural descriptors for neural rendering}. Indeed, with the only exception (VGG metric on Person 1), Deferred Neural Rendering and Neural Point-Based Graphics, which use such learnable descriptors, outperform other methods, sometimes by a rather big margin. 

The relative performance of the two methods that use learnable neural descriptors (ours and DNR) varies across metrics and scenes. Generally, our method performs better on scenes and parts of the scene, where meshing is problematic due to e.g.~thin objects such as 'Studio' (Fig.~\ref{fig:studio-holdout}) and 'Plant' (Fig.~\ref{fig:plant-holdout}) scenes. Conversely, DNR has advantage whenever a good mesh can be reconstructed. 

In support of this observations, user study via Yandex.Toloka web platform was conducted for ScanNet 'Studio' scene and 'Plant' scene. As for 'Studio', we took 300 half image size crops in total uniformly sampled from all holdout images. Labelers were asked to evaluate which picture is closer to a given ground truth crop --- produced by our method or the one produced by DNR. As for 'Plant', 100 random crops of $\frac{1}{6.5}$ original image size were selected. Users have preferred Ours vs. Deferred in \textbf{49.7\% vs. 50.3\%} cases for 'Studio' and in \textbf{69\% vs. 31\%} for 'Plant'. As before, our method performs significantly better when meshing procedure fails in the presence of thin objects and yields results visually similar to DNR when the mesh artefacts are relatively rare.

\begin{figure*}[!h]
\centering
        \begin{overpic}[trim={0 5cm 0 0}, clip, width=.45\textwidth]{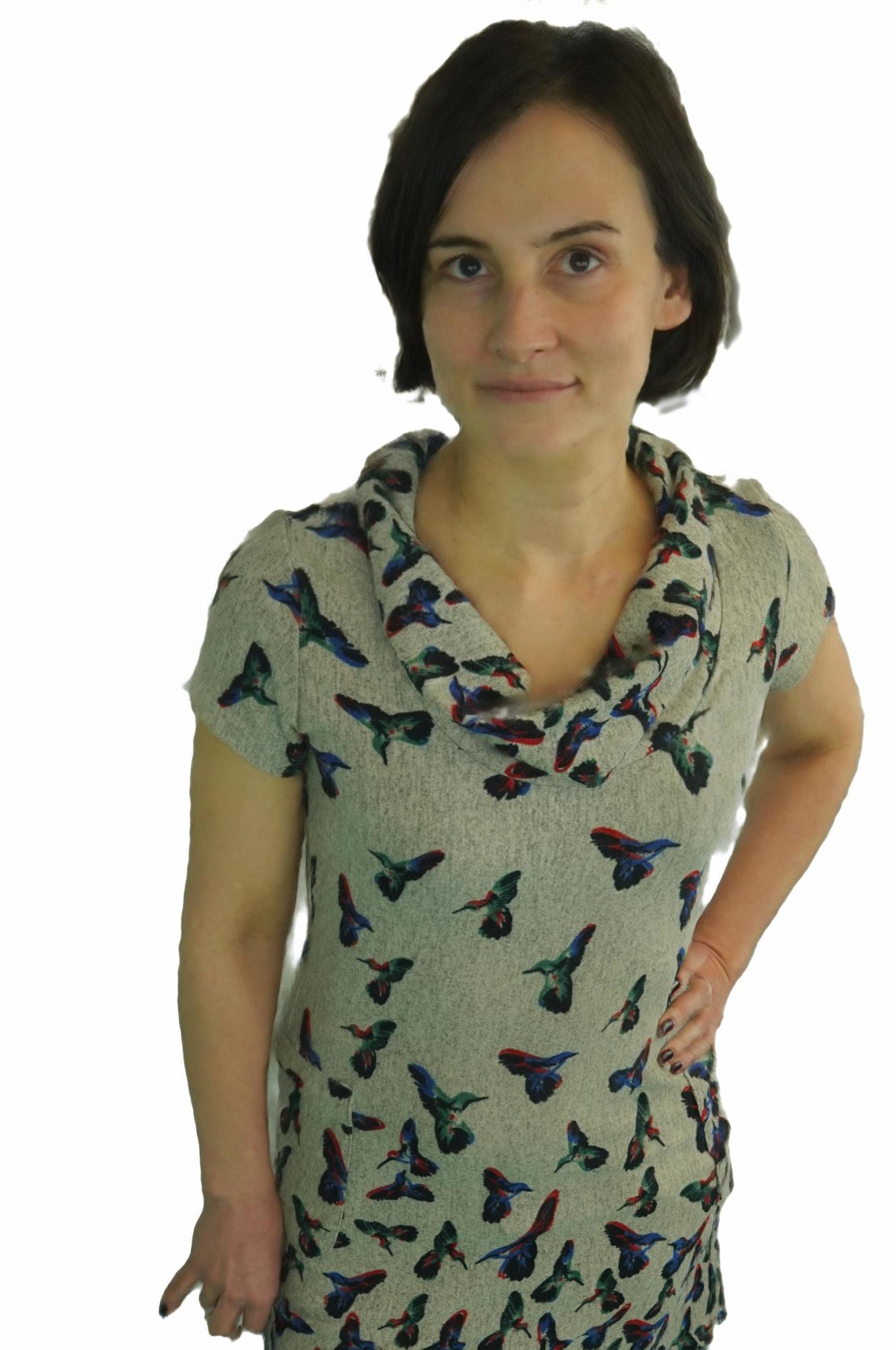}
            \put(-5,-5){\includegraphics[width=.15\textwidth]{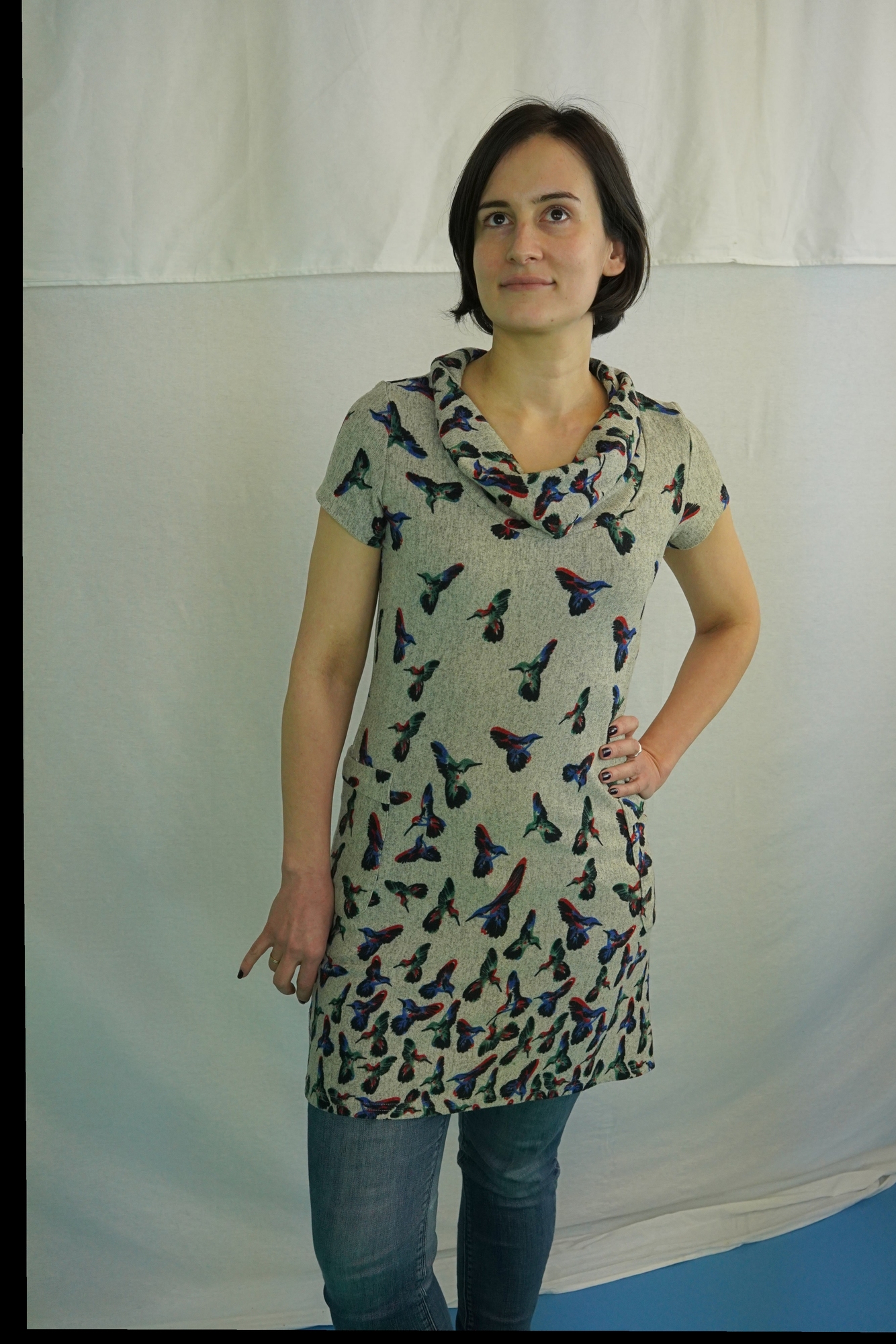}}
        \end{overpic}
        \begin{overpic}[trim={0 5cm 0 0}, clip, width=.45\textwidth]{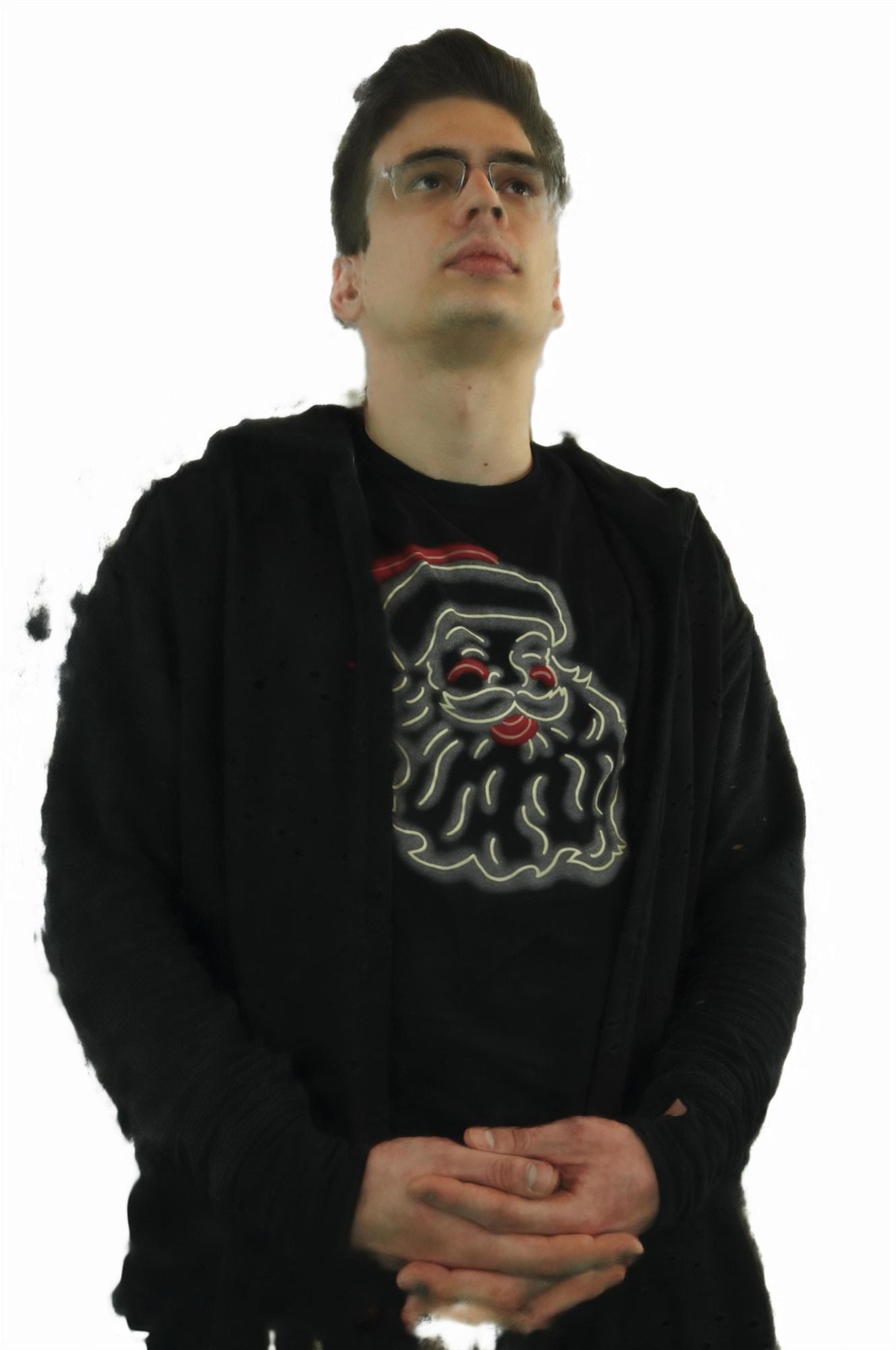}
            \put(-5,-5){\includegraphics[width=.15\textwidth]{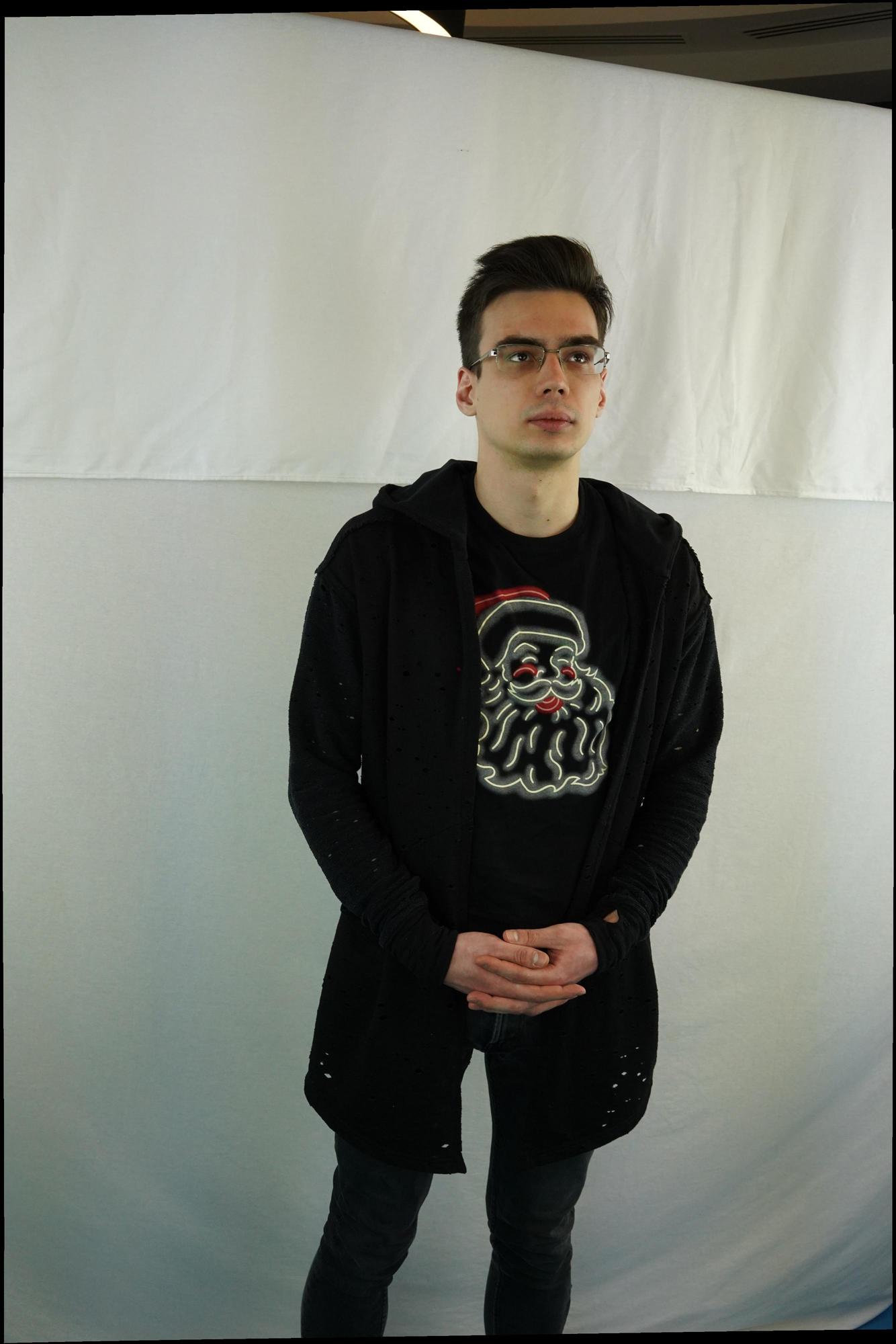}}
        \end{overpic}

    \vspace{0.4cm}
    \caption{Novel views of \textbf{People} generated by our method (\textit{large picture}: our rendered result, \textit{small overlaid picture}: nearest view from the fitting part). Both people are from the holdout part (excluded from pretraining).}
    \label{fig:far_views_people}
\end{figure*}
    
\begin{figure*}[!h]
\centering
    \begin{subfigure}[b]{0.48\linewidth}
        \centering
        \begin{overpic}[width=.49\textwidth]{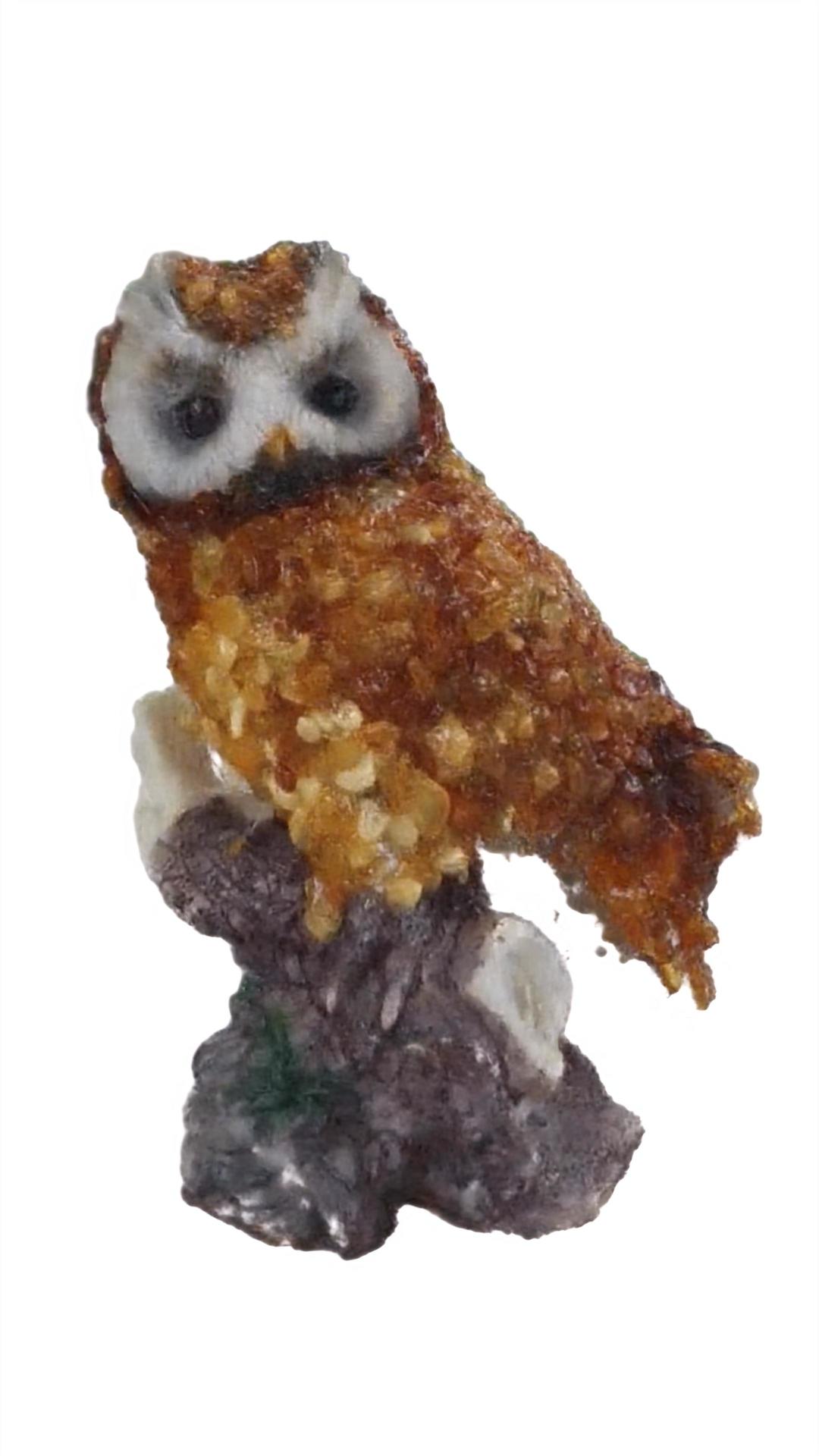}
            \put(0, 0){\includegraphics[trim={11cm 16.5cm 6.5cm 22.5cm}, clip, width=.2\textwidth]{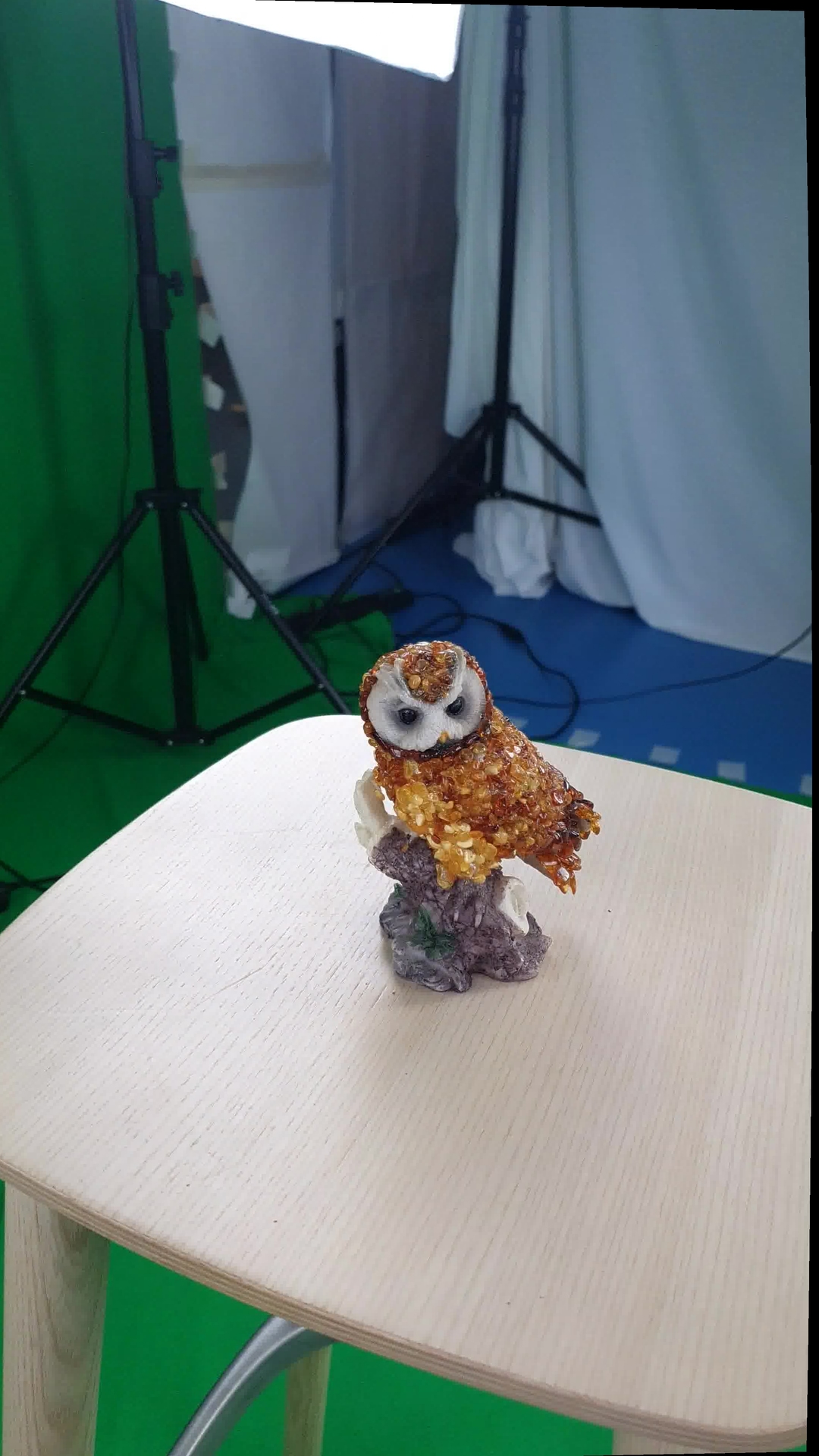}}
        \end{overpic}
        \begin{overpic}[width=.49\textwidth]{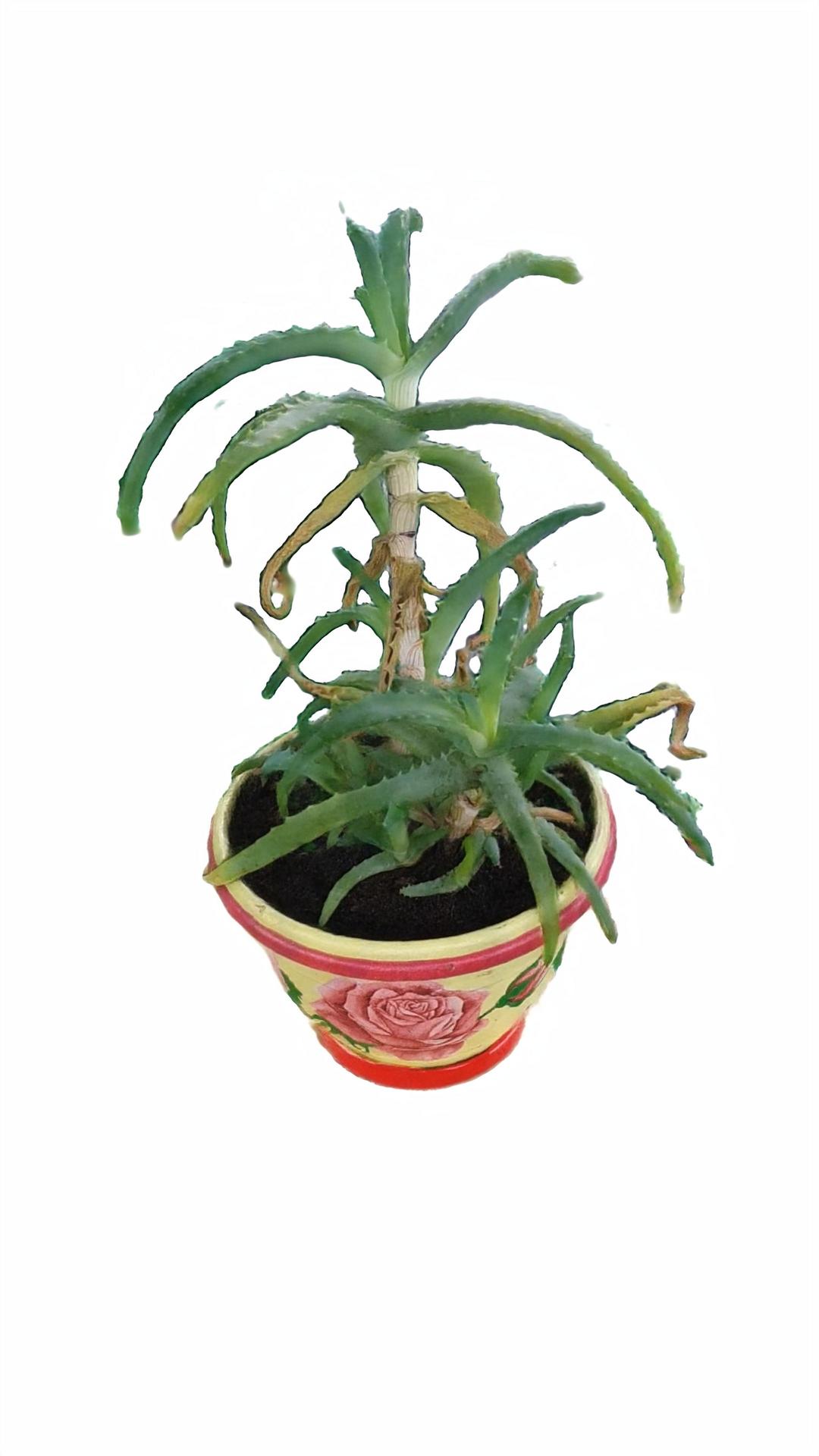}
            \put(0, 0){\includegraphics[trim={6.5cm 16.5cm 11.5cm 22.5cm}, clip, width=.2\textwidth]{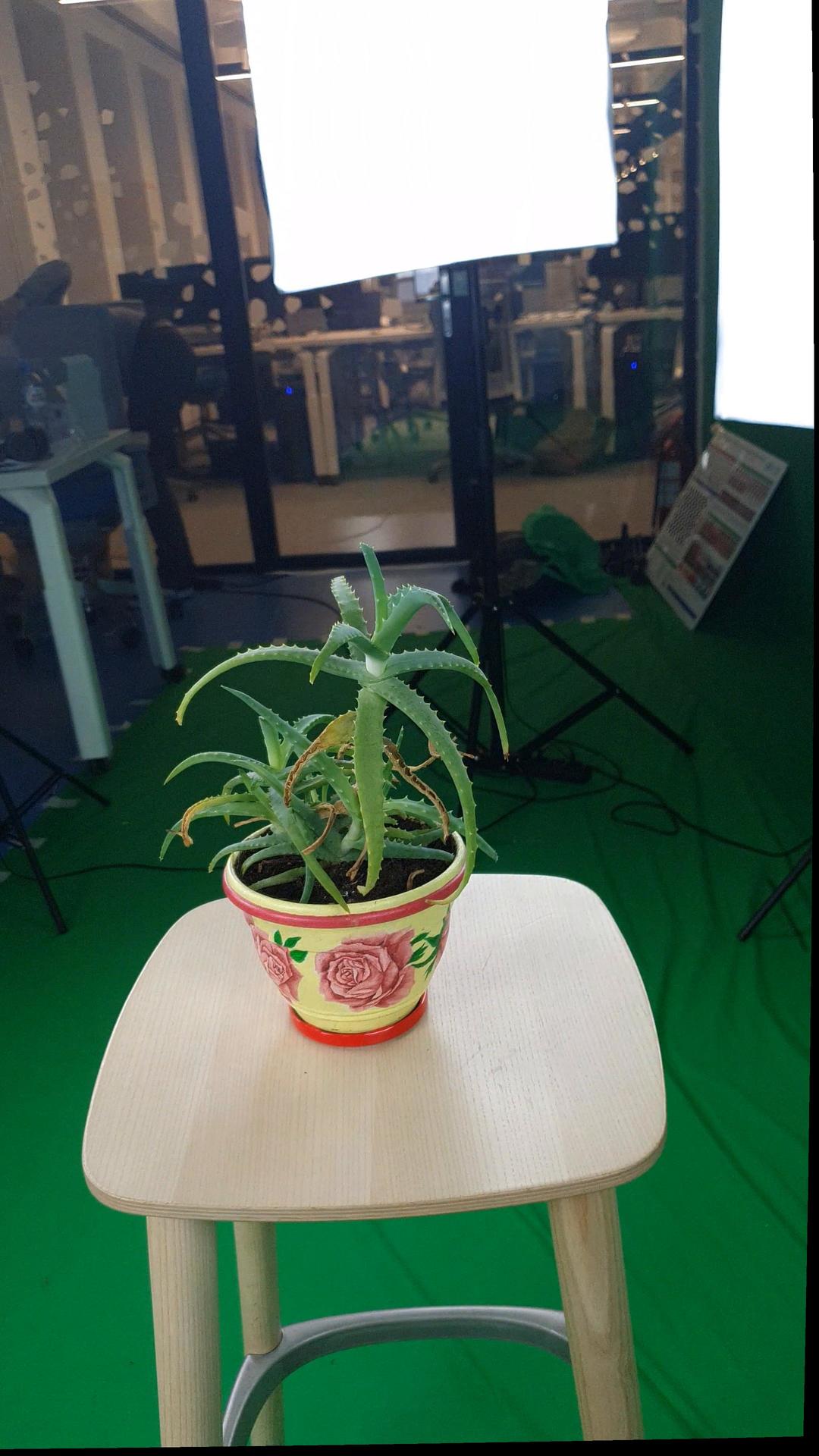}}
        \end{overpic}
        \captionsetup{width=.9\linewidth}
        \caption{\textbf{'Owl'} and \textbf{'Plant'}. \textit{Large picture:} render from a novel point, \textit{small overlaid picture:} nearest train view}
    \end{subfigure}
    \begin{subfigure}[b]{0.48\linewidth}
        \centering
        \begin{subfigure}{0.48\linewidth}
            \includegraphics[trim={0 0 0 0}, clip,width=\textwidth]{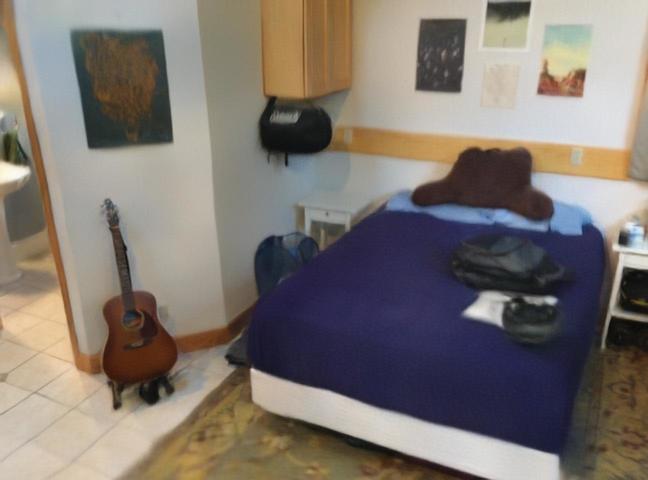}
        \end{subfigure}
        \begin{subfigure}{0.48\linewidth}
            \includegraphics[trim={0 0 0 0}, clip,width=\textwidth]{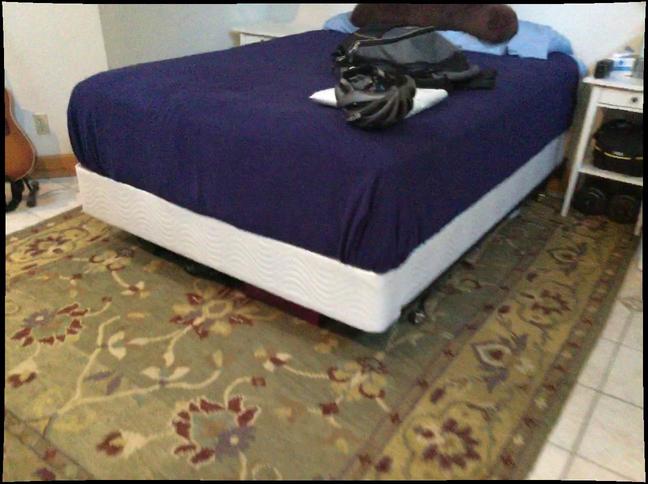}
        \end{subfigure}
        
        \centering
        \begin{subfigure}{0.48\linewidth}
        \includegraphics[trim={0 0 0 0}, clip,width=\textwidth]{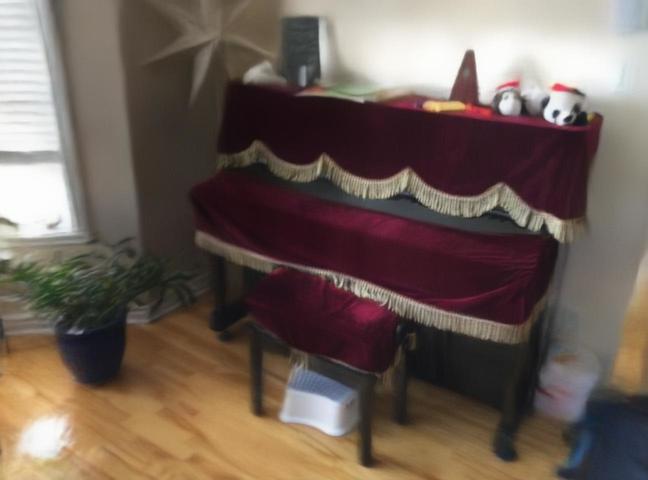}
        \caption*{Ours}
        \end{subfigure}
        \begin{subfigure}{0.48\linewidth}
        \includegraphics[trim={0 0 0 0}, clip,width=\textwidth]{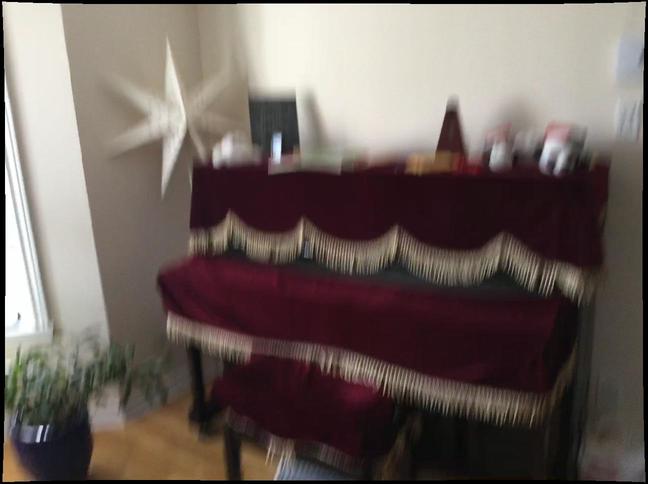}
        \caption*{Nearest train view}
        \end{subfigure}

        \captionsetup[subfigure]{position=b, width=.9\linewidth}
        \caption{\textbf{'Studio'}. \textit{First row:} render from a novel point, \textit{second row:} nearest train view}
    \end{subfigure}
    \caption{Various results obtained by our method. For each of the scenes, we show the view from the nearest camera from the fine-tuning part (\textit{nearest train view}).}
    \label{fig:far_views_things_scannet}
\end{figure*}

\subsubsection{Ablation study.} We also evaluate the effect of some of the design choices inside our method. First, we consider the point cloud density and the descriptor size. We take an evaluation scene from ScanNet and progressively downsample point cloud using voxel downsampling and get small, medium and large variants with 0.5, 1.5, and 10 million points respectively. For each variant of point cloud we fit descriptors with sizes $M$ equal to 4, 8 (default) and 16. \tab{desc_size} shows evaluation results. It naturally reveals the advantage of using denser point clouds, yet it also shows that the performance of our method saturates at $M=8$.

\begin{table}[!h]
    \centering
    \caption{Dependency of the loss function values on the descriptor size and the point cloud size. Comparison is made for the 'Studio' scene of Scannet.}
    \vspace{0.1cm}
    \begin{tabular}{l|c|c|c|c|c|c}
                                     & \multicolumn{2}{c|}{\,Descriptor size 4\,}                                             & \multicolumn{2}{c|}{\,Descriptor size 8\,}                                             & \multicolumn{2}{c}{\,Descriptor size 16\,}                                            \\
    Point Cloud size                 & VGG $\downarrow$    & LPIPS $\downarrow$& VGG  $\downarrow$    & LPIPS  $\downarrow$ & VGG $\downarrow$    & LPIPS $\downarrow$ \\ \hline
    \textit{small}  & 635.74                           & 0.543                           & 632.39                           & 0.505                           & 622.06                           & 0.508                           \\
    \textit{medium} & 622.05                           & 0.506                           & 616.49                           & 0.486                           & 614.90                           & 0.500                           \\
    \textit{large}  & \textbf{610.76} & \textbf{0.509}                           & \textbf{\underline{609.11}} & \textbf{\underline{0.485}}                           & \textbf{611.38} & \textbf{0.488}                         
    \end{tabular}
    \label{tab:desc_size}
\end{table}

\begin{figure*}[!h]
    \centering
    \captionsetup[subfigure]{labelformat=empty, justification=centering}
      \begin{subfigure}{0.31\linewidth}
      \includegraphics[trim={0 22.5cm 7.5cm 0}, clip,width=\linewidth]{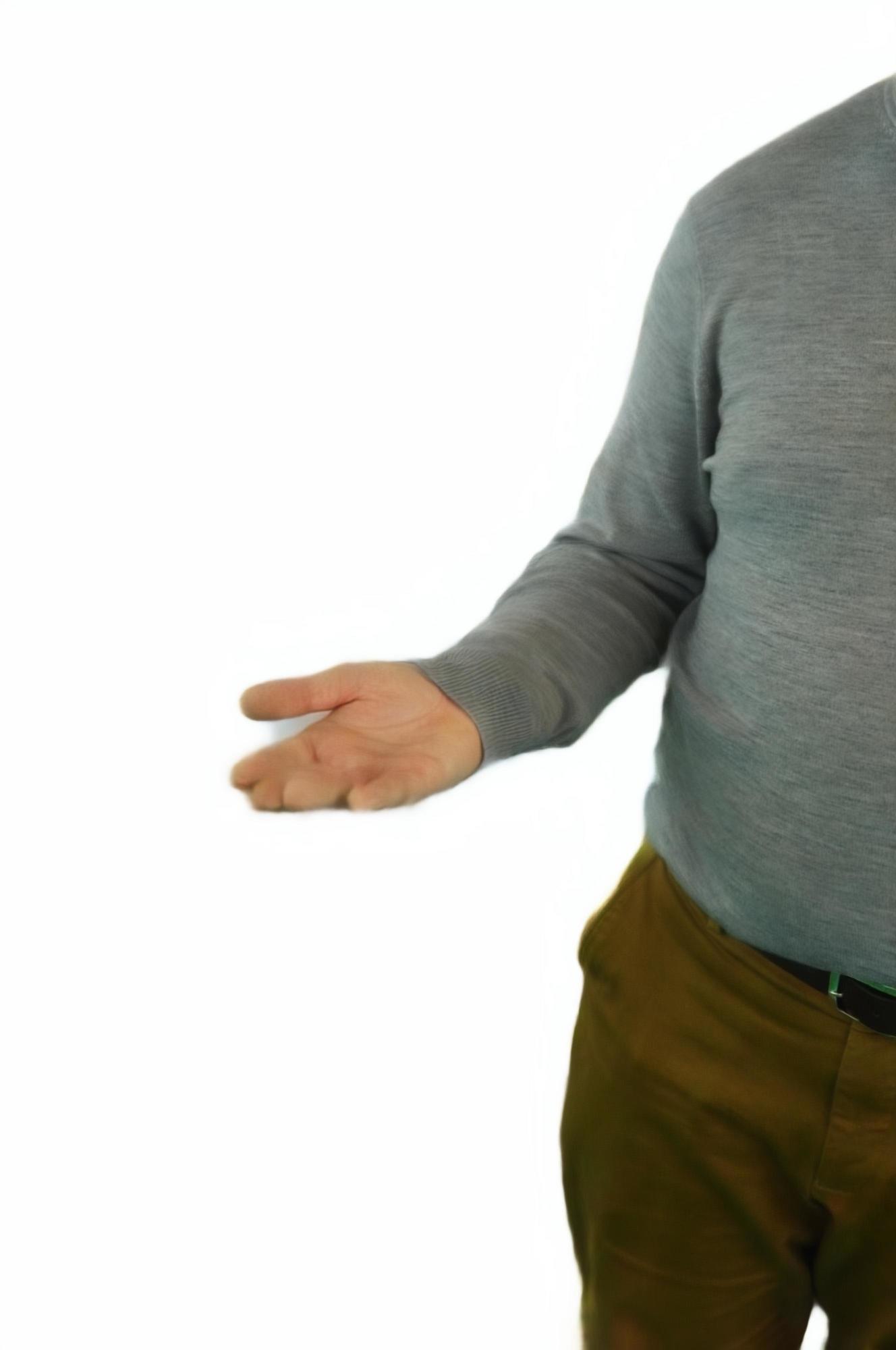}
      \caption{original scene\\ \,}
      \end{subfigure}
      \begin{subfigure}{0.31\linewidth}
      \includegraphics[trim={0 22.5cm 7.5cm 0}, clip, width=\linewidth]{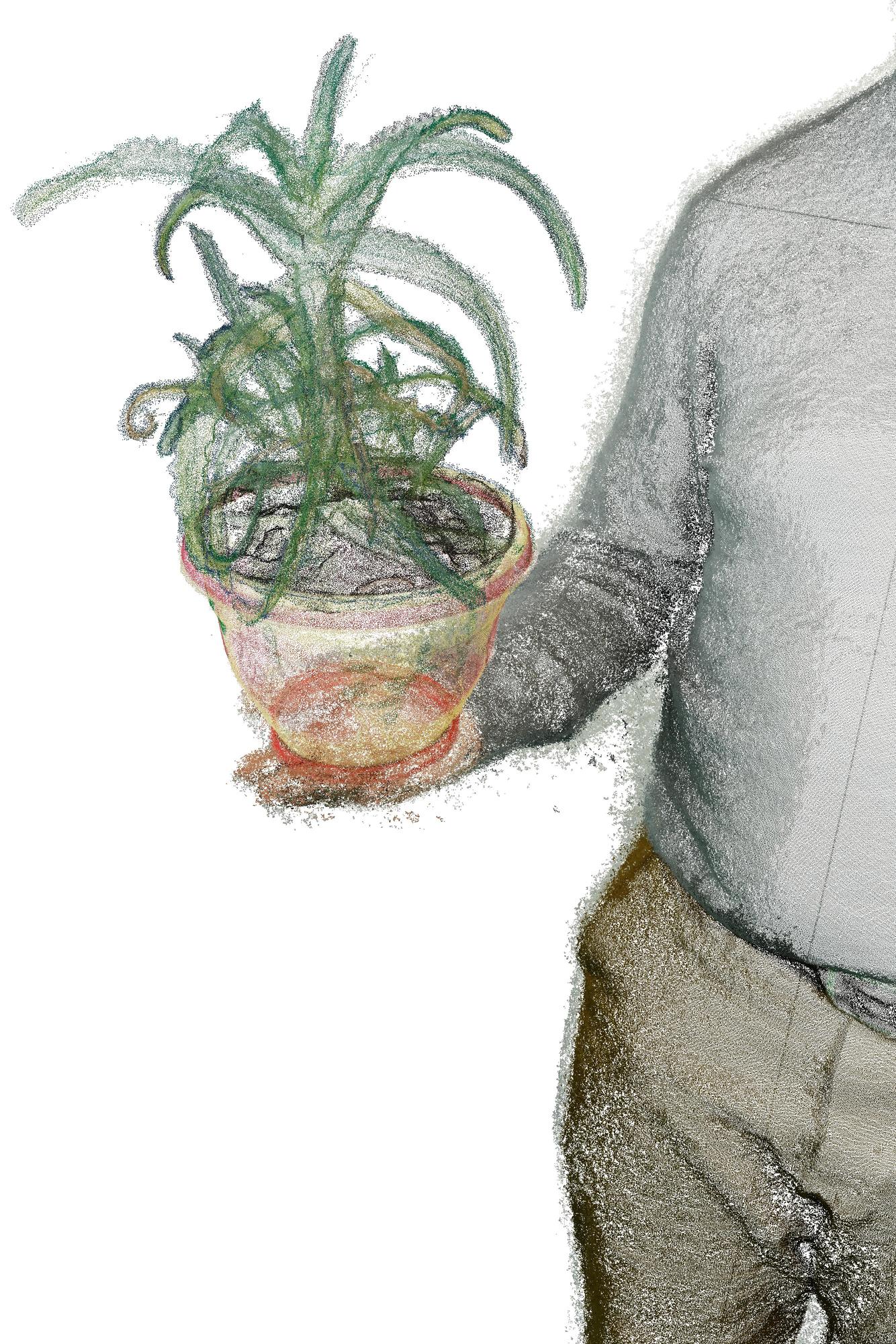}
      \caption{Person 2 + Plant: point cloud composition}
      \end{subfigure}
      \begin{subfigure}{0.31\linewidth}
      \includegraphics[trim={0 22.5cm 7.5cm 0}, clip, width=\linewidth]{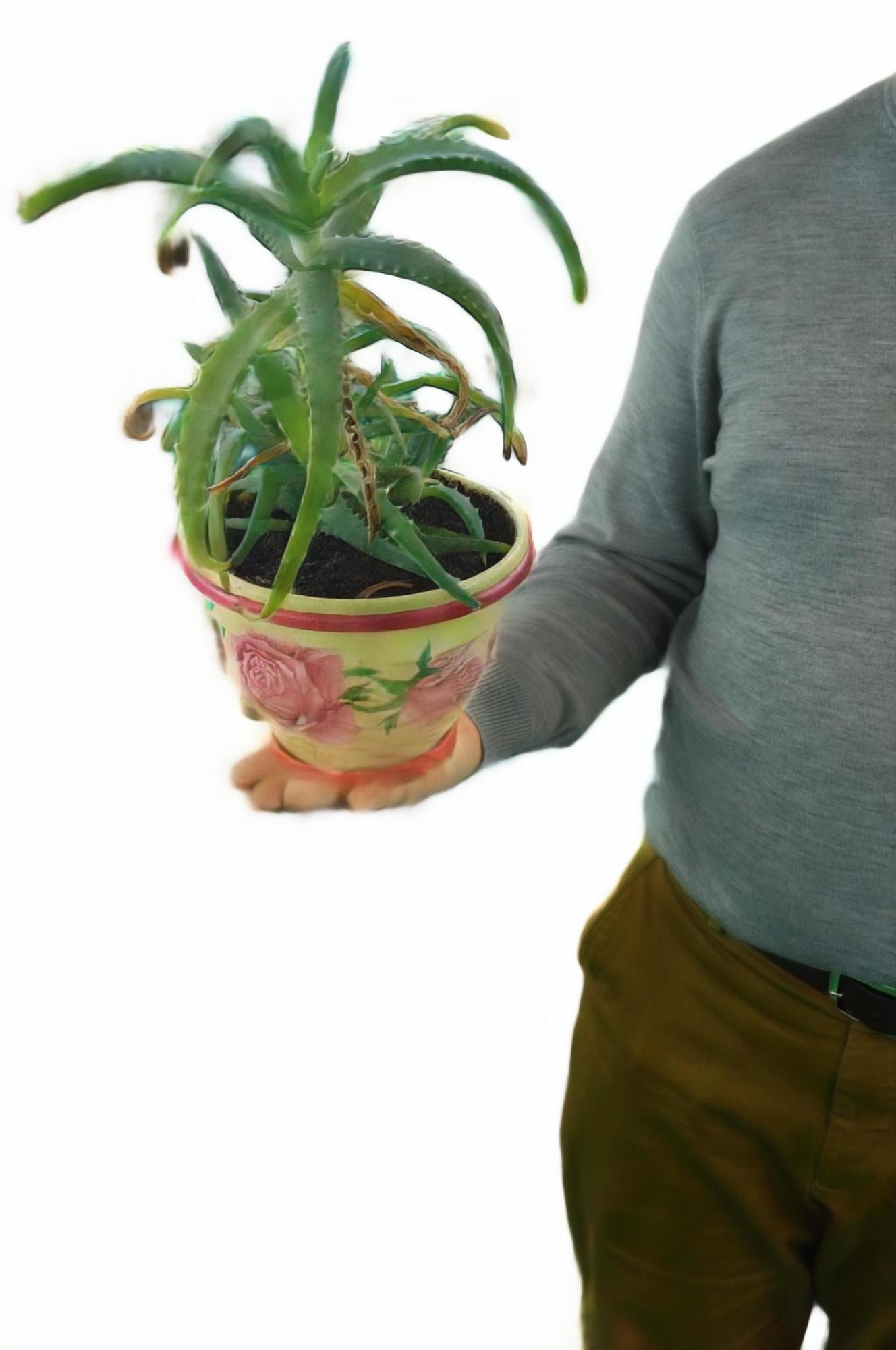}
      \caption{Person 2 + Plant: ours}
      \end{subfigure}
    \caption{A novel view for the composed scenes. A point cloud from 'Plant’ scene was translated and rotated slightly to be placed on the left hand of the 'Person 2'. The world scale of objects was kept unchanged.}
    \label{fig:scene_editing}
\end{figure*}

\subsubsection{Scene editing.} To conclude, we show a qualitative example of creating a composite of two separately captured scenes (Fig.~\ref{fig:scene_editing}). To create it, we took the 'Person 2' and the 'Plant' datasets and fitted descriptors for them while keeping the rendering network, pretrained on People, frozen. We then align the two point clouds with learned descriptors by a manually-chosen rigid transform and created the rendering.

\subsubsection{Anti-aliasing.} We have found that in the presence of camera misregistrations in the fitting set (such as ScanNet scenes), our method tends to produce flickering artefacts during camera motion (as opposed to \cite{Thies19} that tends to produce blurry outputs in these cases). At least part of this flickering can be attributed to rounding of point projections to the nearest integer position during rendering. It is possible to generate each of the raw images at higher resolution (e.g.\ 2$\times$ or 4$\times$ higher), and then downsample it to the target resolution using bilinear interpolation resulting in smoother raw images. This results in less flickering with a cost of barely noticeable blur (see \textbf{supplementary video}). Note that the increase in time complexity from such anti-aliasing is insignificant, since it does not affect the resolution at which neural rendering is performed.

 \section{Discussion}
\label{sect:discussion}

We have presented a neural point-based approach for modeling complex scenes. Similarly to classical point-based approaches, ours uses 3D points as modeling primitives. Each of the points in our approach is associated with a local descriptor containing information about local geometry and appearance. A rendering network that translates point rasterizations into realistic views, while taking the learned descriptors as an input point pseudo-colors. 
We thus demonstrate that point clouds can be successfully used as geometric proxies for neural rendering, while missing information about connectivity as well as geometric noise and holes can be handled by deep rendering networks gracefully. Thus, our method achieves similar rendering quality to mesh-based analog~\cite{Thies19}, surpassing it wherever meshing is problematic (e.g.~thin parts).

{\bf Limitations and further work.} Our model currently cannot fill very big holes in geometry in a realistic way. Such ability is likely to come with additional point cloud processing/inpainting that could potentially be trained jointly with our modeling pipeline. We have also not investigated the performance of the system for dynamic scenes (including both motion and relighting scenarios), where some update mechanism for the neural descriptors of points would need to be introduced.  
\FloatBarrier
\ifnum\value{page}>14 \errmessage{Number of pages exceeded!!!!}\fi

\bibliographystyle{splncs}
\bibliography{refs}

\begin{thebibliography}{10}

\bibitem{Blinn76}
Blinn, J.F., Newell, M.E.:
\newblock Texture and reflection in computer generated images.
\newblock Communications of the ACM \textbf{19}(10) (1976)  542--547

\bibitem{Blinn78}
Blinn, J.F.:
\newblock Simulation of wrinkled surfaces.
\newblock In: Proc. {SIGGRAPH}. Volume~12., ACM (1978)  286--292

\bibitem{Debevec98}
Debevec, P., Yu, Y., Borshukov, G.:
\newblock Efficient view-dependent image-based rendering with projective
  texture-mapping.
\newblock In: Rendering Techniques.
\newblock Springer (1998)  105--116

\bibitem{Wood00}
Wood, D.N., Azuma, D.I., Aldinger, K., Curless, B., Duchamp, T., Salesin, D.H.,
  Stuetzle, W.:
\newblock Surface light fields for 3d photography.
\newblock In: Proc. {SIGGRAPH}. (2000)  287--296

\bibitem{McMillan95}
McMillan, L., Bishop, G.:
\newblock Plenoptic modeling: an image-based rendering system.
\newblock In: {SIGGRAPH}, {ACM} (1995)  39--46

\bibitem{Seitz96}
Seitz, S.M., Dyer, C.R.:
\newblock View morphing.
\newblock In: Proceedings of the 23rd annual conference on Computer graphics
  and interactive techniques, ACM (1996)  21--30

\bibitem{Gortler96}
Gortler, S.J., Grzeszczuk, R., Szeliski, R., Cohen, M.F.:
\newblock The lumigraph.
\newblock In: {SIGGRAPH}, {ACM} (1996)  43--54

\bibitem{Levoy96}
Levoy, M., Hanrahan, P.:
\newblock Light field rendering.
\newblock In: Proceedings of the 23rd annual conference on Computer graphics
  and interactive techniques, ACM (1996)  31--42

\bibitem{Levoy85}
Levoy, M., Whitted, T.:
\newblock The use of points as a display primitive.
\newblock Citeseer (1985)

\bibitem{Grossman98}
Grossman, J.P., Dally, W.J.:
\newblock Point sample rendering.
\newblock In: Rendering Techniques’ 98.
\newblock Springer (1998)  181--192

\bibitem{Gross02}
Gross, M., Pfister, H., Alexa, M., Pauly, M., Stamminger, M., Zwicker, M.:
\newblock Point based computer graphics.
\newblock Eurographics Assoc. (2002)

\bibitem{Kobbelt04}
Kobbelt, L., Botsch, M.:
\newblock A survey of point-based techniques in computer graphics.
\newblock Computers \& Graphics \textbf{28}(6) (2004)  801--814

\bibitem{Isola17}
Isola, P., Zhu, J., Zhou, T., Efros, A.A.:
\newblock Image-to-image translation with conditional adversarial networks.
\newblock In: Proc. {CVPR}. (2017)  5967--5976

\bibitem{Nalbach17}
Nalbach, O., Arabadzhiyska, E., Mehta, D., Seidel, H., Ritschel, T.:
\newblock Deep shading: Convolutional neural networks for screen space shading.
\newblock Comput. Graph. Forum \textbf{36}(4) (2017)  65--78

\bibitem{Chen18}
Chen, A., Wu, M., Zhang, Y., Li, N., Lu, J., Gao, S., Yu, J.:
\newblock Deep surface light fields.
\newblock Proceedings of the ACM on Computer Graphics and Interactive
  Techniques \textbf{1}(1) (2018) ~14

\bibitem{Bui18}
Bui, G., Le, T., Morago, B., Duan, Y.:
\newblock Point-based rendering enhancement via deep learning.
\newblock The Visual Computer \textbf{34}(6-8) (2018)  829--841

\bibitem{Hedman18}
Hedman, P., Philip, J., Price, T., Frahm, J., Drettakis, G., Brostow, G.J.:
\newblock Deep blending for free-viewpoint image-based rendering.
\newblock {ACM} Trans. Graph. \textbf{37}(6) (2018)  257:1--257:15

\bibitem{Pfister00}
Pfister, H., Zwicker, M., Van~Baar, J., Gross, M.:
\newblock Surfels: Surface elements as rendering primitives.
\newblock In: Proceedings of the 27th annual conference on Computer graphics
  and interactive techniques, ACM Press/Addison-Wesley Publishing Co. (2000)
  335--342

\bibitem{Zwicker01}
Zwicker, M., Pfister, H., Van~Baar, J., Gross, M.:
\newblock Surface splatting.
\newblock In: Proc. {SIGGRAPH}, ACM (2001)  371--378

\bibitem{Meshry19}
Meshry, M., Goldman, D.B., Khamis, S., Hoppe, H., Pandey, R., Snavely, N.,
  Martin-Brualla, R.:
\newblock Neural rerendering in the wild.
\newblock In: Proc. {CVPR}. (June 2019)

\bibitem{Pittaluga19}
Pittaluga, F., Koppal, S.J., Kang, S.B., Sinha, S.N.:
\newblock Revealing scenes by inverting structure from motion reconstructions.
\newblock In: Proc. {CVPR}. (June 2019)

\bibitem{Lowe04}
Lowe, D.G.:
\newblock Distinctive image features from scale-invariant keypoints.
\newblock International journal of computer vision \textbf{60}(2) (2004)
  91--110

\bibitem{Flynn16}
Flynn, J., Neulander, I., Philbin, J., Snavely, N.:
\newblock Deepstereo: Learning to predict new views from the world's imagery.
\newblock In: Proceedings of the IEEE Conference on Computer Vision and Pattern
  Recognition. (2016)  5515--5524

\bibitem{Ganin16}
Ganin, Y., Kononenko, D., Sungatullina, D., Lempitsky, V.S.:
\newblock Deepwarp: Photorealistic image resynthesis for gaze manipulation.
\newblock In: Proc. {ECCV}. (2016)  311--326

\bibitem{Zhou16}
Zhou, T., Tulsiani, S., Sun, W., Malik, J., Efros, A.A.:
\newblock View synthesis by appearance flow.
\newblock In: Proc. {ECCV}. (2016)  286--301

\bibitem{Thies18}
Thies, J., Zollh{\"o}fer, M., Theobalt, C., Stamminger, M., Nie{\ss}ner, M.:
\newblock Ignor: Image-guided neural object rendering.
\newblock arXiv 2018 (2018)

\bibitem{Martin18}
Martin-Brualla, R., Pandey, R., Yang, S., Pidlypenskyi, P., Taylor, J.,
  Valentin, J., Khamis, S., Davidson, P., Tkach, A., Lincoln, P.,  et~al.:
\newblock Lookingood: enhancing performance capture with real-time neural
  re-rendering.
\newblock In: SIGGRAPH Asia 2018 Technical Papers, ACM (2018)  255

\bibitem{Sitzmann19}
Sitzmann, V., Thies, J., Heide, F., Nie{\ss}ner, M., Wetzstein, G.,
  Zollh{\"{o}}fer, M.:
\newblock Deepvoxels: Learning persistent 3d feature embeddings.
\newblock In: Proc. {CVPR}. (2019)

\bibitem{Thies19}
Thies, J., Zollh{\"{o}}fer, M., Nie{\ss}ner, M.:
\newblock Deferred neural rendering: Image synthesis using neural textures.
\newblock In: Proc. {SIGGRAPH}. (2019)

\bibitem{Ronneberger15}
Ronneberger, O., Fischer, P., Brox, T.:
\newblock U-net: Convolutional networks for biomedical image segmentation.
\newblock In: International Conference on Medical image computing and
  computer-assisted intervention, Springer (2015)  234--241

\bibitem{Yu18}
Yu, J., Lin, Z., Yang, J., Shen, X., Lu, X., Huang, T.S.:
\newblock Free-form image inpainting with gated convolution.
\newblock arXiv preprint arXiv:1806.03589 (2018)

\bibitem{Williams83}
Williams, L.:
\newblock Pyramidal parametrics.
\newblock In: Proceedings of the 10th annual conference on Computer graphics
  and interactive techniques. (1983)  1--11

\bibitem{Dosovitskiy16}
Dosovitskiy, A., Brox, T.:
\newblock Generating images with perceptual similarity metrics based on deep
  networks.
\newblock In: Proc. {NIPS}. (2016)  658--666

\bibitem{Johnson16}
Johnson, J., Alahi, A., Fei{-}Fei, L.:
\newblock Perceptual losses for real-time style transfer and super-resolution.
\newblock In: Proc. {ECCV}. (2016)  694--711

\bibitem{Simonyan14}
Simonyan, K., Zisserman, A.:
\newblock Very deep convolutional networks for large-scale image recognition.
\newblock CoRR \textbf{abs/1409.1556} (2014)

\bibitem{Diederik14}
Kingma, D.P., Ba, J.:
\newblock Adam: {A} method for stochastic optimization.
\newblock CoRR \textbf{abs/1412.6980} (2014)

\bibitem{Dai17b}
Dai, A., Chang, A.X., Savva, M., Halber, M., Funkhouser, T., Nie{\ss}ner, M.:
\newblock {ScanNet}: Richly-annotated 3d reconstructions of indoor scenes.
\newblock In: Proc. {CVPR}. (2017)

\bibitem{Dai17a}
Dai, A., Nie{\ss}ner, M., Zollh{\"{o}}fer, M., Izadi, S., Theobalt, C.:
\newblock Bundlefusion: Real-time globally consistent 3d reconstruction using
  on-the-fly surface reintegration.
\newblock {ACM} Trans. Graph. \textbf{36}(3) (2017)  24:1--24:18

\bibitem{Gong19}
Gong, K., Gao, Y., Liang, X., Shen, X., Wang, M., Lin, L.:
\newblock Graphonomy: Universal human parsing via graph transfer learning.
\newblock In: Proceedings of the IEEE Conference on Computer Vision and Pattern
  Recognition. (2019)  7450--7459

\bibitem{agisoft}
{Agisoft}:
\newblock Metashape software. (retrieved 20.05.2019)

\bibitem{Zhang18}
Zhang, R., Isola, P., Efros, A.A., Shechtman, E., Wang, O.:
\newblock The unreasonable effectiveness of deep features as a perceptual
  metric.
\newblock In: CVPR. (2018)

\bibitem{Heusel17}
Heusel, M., Ramsauer, H., Unterthiner, T., Nessler, B., Hochreiter, S.:
\newblock Gans trained by a two time-scale update rule converge to a local nash
  equilibrium.
\newblock In: Advances in neural information processing systems. (2017)
  6626--6637

\end{thebibliography}
\end{document}